\documentclass{article}
\usepackage{spconf,amsmath,graphicx}
\usepackage{hyperref}
\hypersetup{hidelinks}
\usepackage{graphicx}
\usepackage{float} 
\usepackage{enumerate}
\usepackage{verbatim}
\usepackage{svg}
\usepackage{tabularx}
\usepackage{amssymb}
\usepackage{multirow}
\usepackage{multicol}
\usepackage{CJKutf8}
\usepackage{caption}
\usepackage{pbox}
\usepackage{array}
\usepackage{booktabs}
\usepackage{color}
\usepackage{longtable}

\def\x{{\mathbf x}}
\def\L{{\cal L}}
\makeatletter
\def\UrlAlphabet{%
      \do\a\do\b\do\c\do\d\do\e\do\f\do\g\do\h\do\i\do\j%
      \do\k\do\l\do\m\do\n\do\o\do\p\do\q\do\r\do\s\do\t%
      \do\u\do\v\do\w\do\x\do\y\do\z\do\A\do\B\do\C\do\D%
      \do\E\do\F\do\G\do\H\do\I\do\J\do\K\do\L\do\M\do\N%
      \do\O\do\P\do\Q\do\R\do\S\do\T\do\U\do\V\do\W\do\X%
      \do\Y\do\Z}
\def\UrlDigits{\do\1\do\2\do\3\do\4\do\5\do\6\do\7\do\8\do\9\do\0}
\g@addto@macro{\UrlBreaks}{\UrlOrds}
\g@addto@macro{\UrlBreaks}{\UrlAlphabet}
\g@addto@macro{\UrlBreaks}{\UrlDigits}
\makeatother
\begin{document}
\topmargin=0mm

\title{IFDID: information filter upon diversity-improved decoding for diversity-faithfulness tradeoff in NLG}
%
\name{Han Meng$^{\ddagger}$\sthanks{Work done during an internship at OPPO Research Institute.} \qquad Xiaosong He$^{\heartsuit}$ \qquad Zexing Chen$^{\clubsuit, \spadesuit}$ \qquad Feng Zhou$^{\heartsuit}$\sthanks{Corresponding author, zhoufeng1@oppo.com}}

\address{$^{\ddagger}$ Beijing University of Post and Telecommunications, School of Computer Science, Beijing, China \\
$^{\heartsuit}$ OPPO Research Institute, Beijing, China\\
$^{\clubsuit}$ University of Electronic Science and Technology of China, UoG-UESTC Joint School, Chengdu, China\\
$^{\spadesuit}$ University of Glasgow, Glasgow, United Kingdom}
%
%
%

%
\maketitle
\begin{abstract}
Some Natural Language Generation (NLG) tasks require for both faithfulness and diversity. The decoding strategy is intensively related to the quality of generated text. Strategies such as beam search, greedy search, etc. perform with low diversity and high repetition. On the other hand, guided decoding, the solution towards diversity, may cause unfaithful generation. To this end, this paper presents \textbf{I}nformation \textbf{F}ilter upon \textbf{D}iversity-\textbf{I}mproved \textbf{D}ecoding (\textbf{IFDID}) to obtain the tradeoff between diversity and faithfulness. IFDID is a two-stage decoding strategy leveraging Enhance-Filter framework, for which achieves the tradeoff by increasing the probability of some typical tokens and subsequently filters them by their information amount. To verify the effectiveness of the proposed method, we compare our method with other baselines on related CommonGEN, RocStories and AdGen benchmarks, which cover Chinese and English datasets. Our numerical experimental results and human evaluation outcomes verify the effectiveness of the proposed approach, as our approach achieves 1.24 higher ROUGE score describing faithfulness as well as higher diversity represented by 62.5\% higher upon Dist-2 than traditional approaches, demonstrating that IFDID is a novel SOTA decoding strategy for tradeoff between diversity and faithfulness\footnote{Our code is available at \url{https://github.com/NILIKUO/IFDID}.}.
\end{abstract}
\section{Introduction}
\label{sec:intro}

\begin{figure}[ht]
\centering
\includegraphics[width=0.49\textwidth]{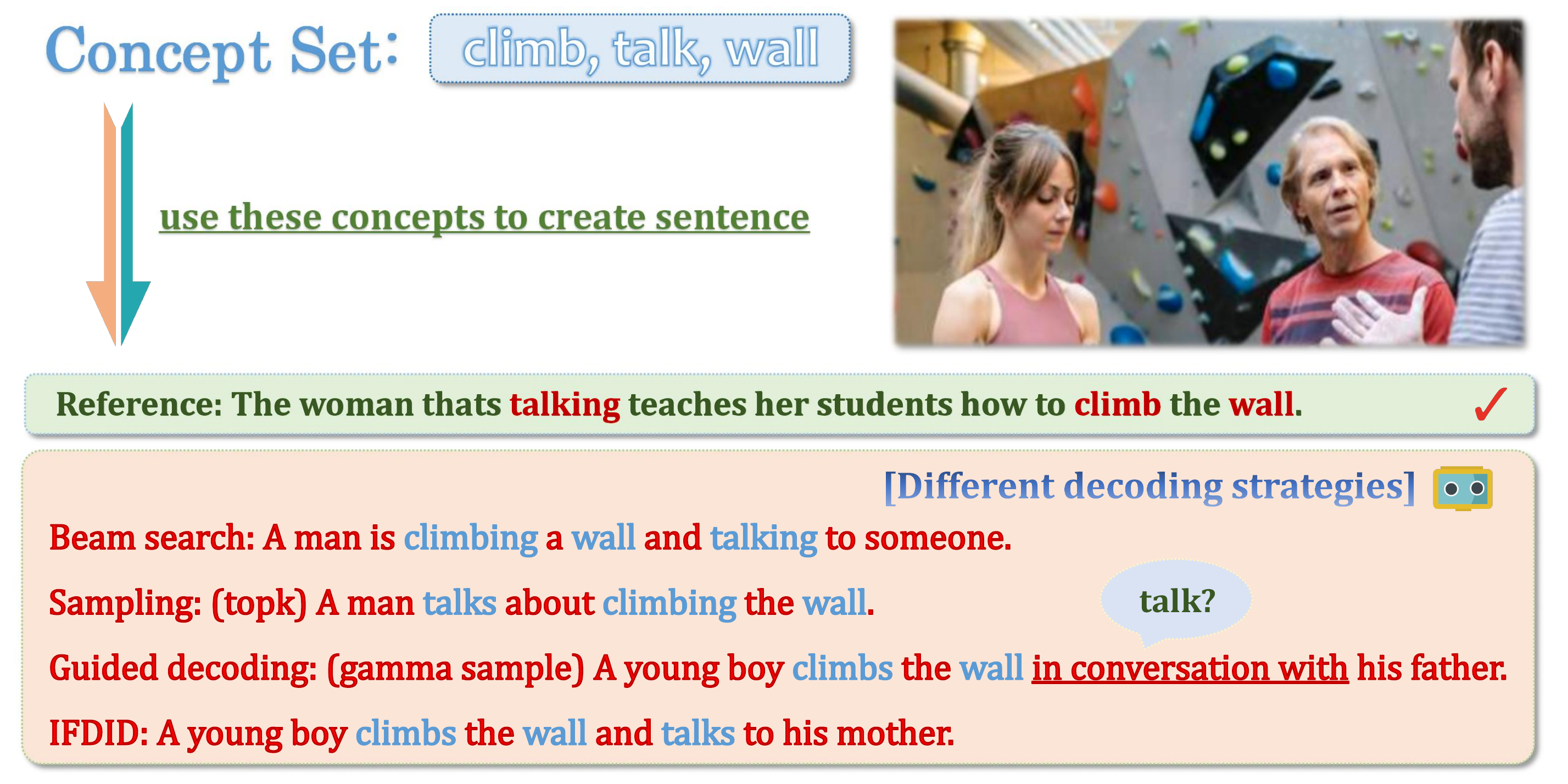}
\caption{Features of beam search, sampling, guided decoding and proposed decoding strategies in CommonGEN \cite{commongen} task. The output of beam search and sampling method is lack of diversity but faithful, whereas result generated by guided decoding strategies is diverse but fail to cover all information in concept set. IFDID, the proposed approach, successfully strike balance between faithfulness and diversity.}
\label{fig:fig1}
\end{figure}

Natural Language Generation (NLG) \cite{nlgsurvey}, consisting of text summarization \cite{sum, sum2}, machine translation \cite{Hutchins1992AnIT}, question generation \cite{zhou2017neural}, etc., is perceived as one of the substantially vital tasks in Natural Language Processing (NLP) and have received great attention recently. The primary objective of NLG is to generate high-quality\footnote{We generally assume 'high-quality' text refers to the text with high score based on evaluation mentioned in \cite{evaluation}, like coherent, diverse, etc.} text based on the input data. However, there is considerable dichotomy lay in evaluation metrics of 'high-quality' among specific tasks. For instance, in summarization task, the evaluation puts heavy emphasis on faithfulness, i.e., whether the generated text can precisely summarize the given source text. In contrast, the diversity of outputs will not be emphasized here \cite{composition}. Similarly, the quality of machine translation result will likewise not be evaluated on the diversity metric \cite{bleu}. Nevertheless, another example is story generation \cite{story} task, which is an open-domain task. In this case, the diversity and interestingness \cite{astar} are usually highly regarded, therefore the diversity metric becomes a decisive criteria. Thus, it can be concluded that the evaluation metric of various tasks in NLG may distinct from the others significantly.

However, there are a series of tasks in NLG that value both diversity and faithfulness. For example, in the data-to-text task \cite{data2text}, paraphrase generation \cite{paraphrase} task, as well as some downstream tasks such as product description generation in e-commerce \cite{eCommerce} and text deduplication \cite{deduplication}, it is necessary to accurately cover the content in the source text whilst avoiding generate dull, awkward as well as repetitive text at the same time. In these cases, the evaluation metric focuses less on the faithfulness but more on diversity while comparing with tasks such as machine translation and summarization generation. Hence, the faithfulness-diversity tradeoff became a major challenge of this set of tasks.

For these tasks that equally emphasize on diversity and faithfulness, it is extraordinary challenging to generate high-quality text. The quality of generated text is intensively relevant to decoding strategies. {\it beam search} \cite{beam}, expanding the most promising node in a small set to explore a graph, is one of the decoding strategies that frequently-used, which is based on heuristic search. It would sort vocabularies by its probabilities through some heuristic methods and select the high-probability phrases to generate sentences. Its tenet is bound to result in its terrible performance on diversity metric, and degeneration \cite{topp, unlikelihood}, which is leveraged to delineate the phenomenon of generating incoherent, unrelated and repetitive text, might even appear in some certain tasks. Subsequently, the {\it sampling} method could somehow alleviate degenerating with its multifarious parameters combination. However, in order to balance faithfulness, it struggles to generate high quality text and is still in lack of diversity and failed to present copious expressions in syntax and sentence patterns. On the other hand, to further improve the diversity, a wide range of decoding strategies, a.k.a guided decoding \cite{astar, ctrl, guided}, such as gamma sample \cite{gamma}, diverse beam search \cite{diverse}, etc. are proposed. These decoding strategies saliently ameliorate the diversity of generated text, but resulted in unfaithful outputs. The features of above-mentioned different decoding strategies are demonstrated in Fig. \ref{fig:fig1}. It remains unclear that how to strike a balance between faithfulness and diversity.

To address these shortcomings, namely the tradeoff between diversity and faithfulness, this paper proposes a new decoding strategy, {\bf IFDID}, namely {\bf I}nformation {\bf F}ilter upon {\bf D}iversity-{\bf I}mproved {\bf D}ecoding, focusing on the balance between diversity and faithfulness, which is leveraged in the decoding stage. IFDID could be considered as two-stage decoding strategy utilizing {\bf Enhance-Filter} framework. In {\bf Enhance} stage, the probabilities of some typical tokens of the source text will be incremented in order to enhance the diversity. The Enhance stage is based on sampling method. Whilst probabilities of more tokens are mounted, the random selection range is enlarged, leading to diverse choice of tokens. The Enhance is done by a function between pre-probabilities and post-probabilities, and the function could be various as well as flexible. A brand new map function, similarity-based function, is proposed, while an add-on decoding strategy, {\bf IFDID-SIMI}, a.k.a {\bf IFDID-SIMI}larity, is presented as well. W.r.t IFDID-SIMI, on the premise of IFDID, the probability modification of each token could be distinctive from each other decided by their word embedding's similarity, which is a novel map function between probabilities. The probability of more decisive token being selected will reap more promotion, and vice versa. The general Enhance function and IFDID-SIMI function could be perceived as two practical implementation of the map function.
Afterwards, inspired by the hypothesis of text that is seen as human-like should have information that is close to the entropy, or anticipated information, of natural language strings \cite{information}, {\bf Filter} stage could be perceived as an entropy filter aiming to augment and ensure faithfulness. Further, similar idea is embodied in nucleus sampling \cite{topp}, which is a probability filter in decoding stage.

Extensive experiments on three generation tasks, namely commonsense generation (CommonGEN \cite{commongen}), story generation (ROCStories \cite{rocstories}) and text generation with specific style (AdGen \cite{adgen}) are carried out and demonstrate that our approach outperforms competitive benchmarks along several metrics. We evaluate faithfulness and diversity by automatic evaluation and human evaluation. Intriguingly, we observe that, compared to beam search and sampling methods: (a) IFDID proffers more diversity, and (b) text generated from beam search and sampling methods upon some tasks may beget degeneration, about which a prospective theoretic explanation is given in paper. Against guided decoding strategies represented by gamma sample, (a) IFDID could mitigate the sacrifice of faithfulness while promoting diversity as well as (b) IFDID-SIMI could enhance diversity under the condition of slight reduction of faithfulness.

In summary, our key contributions are as follows.
\begin{itemize}
\item We propose a two-stage decoding strategy, IFDID, which could perform well and generate high-quality text in any NLG task that demands both diversity and faithfulness leveraging an Enhance-Filter framework.
\item We extend the Enhance stage of IFDID as well as a novel, pilot similarity-based map function, IFDID-SIMI, is presented, which could flexibly control the probabilities of tokens by word embedding's similarity.
\item A theoretic hypothesis on degenerations is proposed and the circumstances that degeneration would happen are empirically discussed with specific case study.
\item We compare IFDID widely with {\it beam search}, sampling methods including {\it top-k sampling}, {\it nucleus sampling}, etc. and guided decoding methods on several popular benchmarks, demonstrating the effectiveness of our method across English and Chinese. Experiments show that proposed approach could achieve tradeoff between faithfulness and diversity.
\end{itemize}

\section{Related Work}
\label{sec:format}

\subsection{Diversity-oriented approaches for NLG}
In this section, we describe a series of proposed approaches to address the lack of diversity caused by the degeneration phenomenon. {\bf Model-based improvement} and {\bf decoding-based improvement} are the main topics of earlier efforts. These improvement techniques are crucial and regularly suggested, especially in data-to-text \cite{data2text} such as graph-to-text \cite{webnlg}, keyword-to-text \cite{key}, SQL-to-text \cite{sql} as well as table-to-text \cite{table}, and paraphrase \cite{paraphrase} tasks that call for faithfulness and diversity.\footnote{To summarize it, conditional text generation and constrained text generation task more inclined to prefer both faithfulness and diversity, and the task definition refers to \cite{constrained}.}
\label{ssec:subhead}

{\bf Model-based approaches.} For the researches on model-based measures, \cite{coverage} utilized coverage loss, \cite{simctg} suggested contrastive loss, whereas \cite{CTLoss} proposed contrastive token objective, a.k.a CTloss, for each of which to discourage repetition. Apart from this, \cite{unlikelihood} put forward a new unlikelihood objective and vindicated that token as well as sequence level unlikelihood training have effect on diminishing repetition. \cite{ctrl}, on the other hand, trained the model with control codes, enabling users to explicitly increase the diversity of generated text.

{\bf Decoding-based approaches.} As for the investigations on decoding-based methods, \cite{topp} proposed nucleus sampling, which randomly select tokens among those containing the vast majority of probability mass. Similarly, the idea in \cite{gamma} suggested to increase the probability to choose attribute-related tokens, including synonyms, to enlarge the scope of possible tokens. Besides, the composition sampling provided by \cite{composition} utilized the entity chain to sample compositions and generate diverse meaningful outputs. By grouping the beam search budget and imposing dissimilarity between beam groups, diverse beam search (DBS) \cite{diverse} produces a list of disparate outputs in order to exalt diversity. The techniques given by \cite{forbidden-ngram} is also developed on the basis of beam search, which proposed the idea of avoid duplicate n-grams and would be an effective measure in promoting diversity as well.

\subsection{Discussion of tradeoff principle}

\begin{figure}[ht]
\centering
\includegraphics[width=0.49\textwidth]{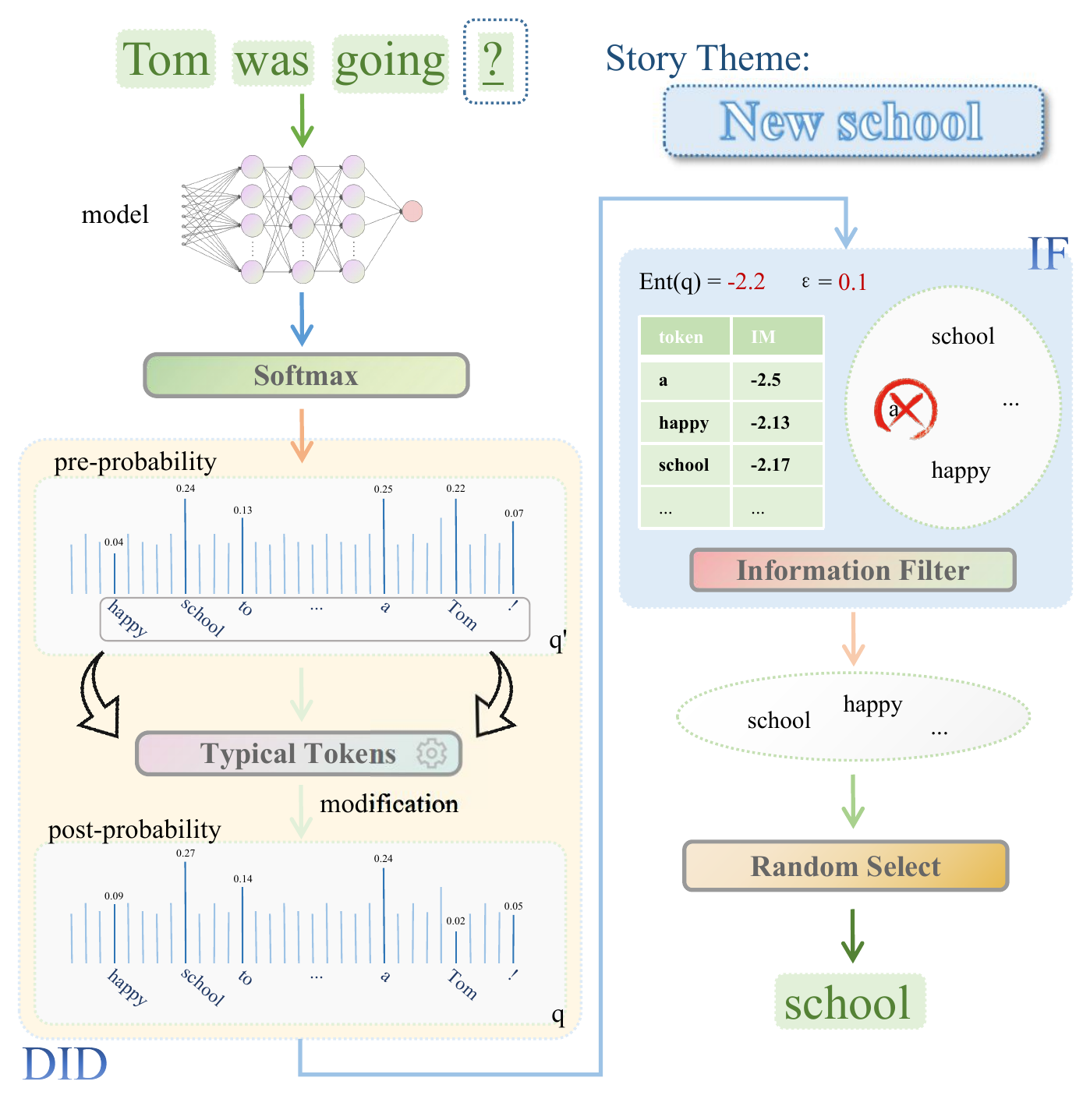}
\caption{Overview of IFDID.}
\label{fig:fig2}
\end{figure}

The concept of tradeoff in natural language generation was firstly proposed in \cite{quality-diversity-tradeoff}, which emphasized on the tradeoff between quality and diversity, then evaluated several decoding strategies on these two scales. \cite{strategy-task} also performed the evaluation of divergent decoding methods across varied tasks, verified the quality-diversity tradeoff and revealed the fact that decoding strategies were usually optimized for specific tasks. The idea of tradeoff was also elaborated by \cite{oversampling}, where diversity was improved by oversampling additional candidates in order to dwindle the sacrifice of faithfulness.

Additionally, a new perspective from the information theory is proposed. Previous work demonstrated that a novel manner of perceiving uniform information density (UID) as a regularizer for training models could improve perplexity \cite{uid-cognitive} as well as \cite{uid-beam} pioneering suggested beam search that enforced UID in text. Subsequently, the expected information hypothesis unveiled that high-quality text had entropy around the entropy of natural strings. \cite{information}

\section{Methodology}
\label{sec:pagestyle}
\subsection{Foundation: Gamma sample}
\label{sec:foundation}
{\bf Decoding.} The process of creating an output sequence $\boldsymbol{y}$ from an input sequence $\boldsymbol{x}$ is known as sequence-to-sequence generation. We consider the generation model:

\begin{equation}
p_\theta(\boldsymbol{y}|\boldsymbol{x})=\prod_{t=1}^{|\boldsymbol{y}|}p_\theta(y_t|\boldsymbol{y}_{<t},\boldsymbol{x}), \label{eq1}
\end{equation}

where $\theta$ are the model parameter set. Assuming that the generative model is an autoregressive, unidirectional model, the decoding strategy is used to select a token from vocabulary set after obtaining the distribution of the next token upon vocabulary set. A decoding algorithm could be perceived as a function $\boldsymbol{f}$:

\begin{equation}
\boldsymbol{idx}=\boldsymbol{f}(\boldsymbol{q}(y_t)), \label{eq2}
\end{equation}

where $\boldsymbol{q}$ is probability distribution of $\boldsymbol{t}$-th token, $\boldsymbol{idx}$ represents a token selected as the next token of the sentence. Decoding strategy is aiming at selecting the best token at each step, in order to improve text quality.

{\bf Gamma sample} \cite{gamma} is a decoding algorithm based on idea that the generated text quality is intensively related to some typical tokens. Hence, controlling the probabilities of these typical tokens is substantially crucial and is done by a gamma parameter inspiring by gamma correction\footnote{\url{https://en.wikipedia.org/wiki/Gamma_correction}}:

\begin{equation}
p^*=g(\gamma,p), \label{eq3}
\end{equation}

where $p^*$ is the subsequent probability distribution of typical tokens, as well as function $g$ is the process of gamma sample and probabilities of typical tokens are monotonically increased. Specifically, the way to change these typical tokens is maintaining the set of typical tokens of the current decoding step, $\mathcal{T}$, and the set of frozen tokens, $\mathcal{F}$, whose probabilities have been modified in previous steps. The computational approach of $g$ is:

\begin{equation}
\begin{aligned}
p^*_\mathcal{T}&=p^{\tan(\frac{\pi \gamma}{2})}_\mathcal{T}\cdot (1-p_\mathcal{F})^{1-\tan (\frac{\pi \gamma}{2})}, \\
p_t^*&=p_t\cdot \frac{p_\mathcal{T}^*}{p_\mathcal{T}},\quad t \in \mathcal{T},\\
p_i^*&=p_i+\frac{p_\mathcal{T}-p_\mathcal{T}^*}{p_{\complement_{\mathcal{F}\cup \mathcal{T}}}},\quad i\notin \mathcal{F}\cup\mathcal{T},
\end{aligned}
\label{eq4}
\end{equation}

where $h(\gamma)=\tan (\frac{\pi \gamma}{2})$ is activation function, $p_\mathcal{T}$ refers to the sum of probabilities of tokens in $\mathcal{T}$, while $p_\mathcal{F}$ is the total of probabilities of tokens in $\mathcal{F}$, similarly, $p_{\complement_{\mathcal{F}\cup \mathcal{T}}}$ refers to the total of probabilities of tokens neither in $\mathcal{T}$ nor $\mathcal{F}$. The purpose of controlling the quality of generated text is achieved by modifying the probability of these typical tokens.

\subsection{Diversity-improved approach}
\label{sec:did}
Based on the foundation of gamma sample, the design of typical tokens set is immensely significant. We propose a novel, effective typical tokens set construction approach.

Formally, we define the input is a set of input pieces $\boldsymbol{I}=\{x_1,x_2,...,x_n\}$, and output is $\boldsymbol{O}=\{y_1,y_2,...,y_m\}$, where $x_i$, $y_i$ represents a single token. The typical tokens selection principle is demonstrated below.

{\bf Theme tokens.} This set of tokens is selected in order to control the relatedness of generated text, which is designed as a list of $x_i$.

{\bf Terminal tokens.} Within this set of tokens, the sentence length could be governed well. The punctuation such as '.', '!', '?', etc., which frequently appears at the end of a sentence should be picked.

{\bf Repeated tokens.} The main issues that hinder diversity enhancement include duplication, which could lead to degeneration phenomenon \cite{topp}. In order to boost diversity, tokens that have appeared in the previous output sequence should be penalized. We collect these tokens into repeated tokens set.

Leveraging the novel typical tokens set construction approach and backbone of gamma transformation in Sec. \ref{sec:foundation}, more typical tokens probabilities will be modified, and the diversity of generated text will be improved. The following are some of the novel decoding principles proposed in this paper.

{\bf When meeting with extremeness.} During decoding steps, some extreme conditions may occur, which refers to the probability of a certain token approximately equal to 0 or 1. Furthermore, the probability distribution is uneven in certain phases, and some extreme situations arise. Since this condition is common in some decoding steps, ignoring these situations is not an option. In other words, keeping these circumstances is the best approach that this paper employs. If the probability distribution obtained in this decoding step contains a token whose probability is closer to 0 or 1 and goes beyond a certain threshold, such as $10^{-6}$, then modify the probability of this token to the value that is closest to 0 or 1 and does not exceed the threshold. Lastly, normalization is necessary after the modification.

{\bf When facing other languages that leading to some key information appear in one input piece.} Language like Chinese allows multiple tokens to form a word and each token has a distinctive meaning. For instance, \begin{CJK}{UTF8}{gbsn}'概率'\end{CJK} means 'probability', but \begin{CJK}{UTF8}{gbsn}'概'\end{CJK} means 'concept' as well as \begin{CJK}{UTF8}{gbsn}'率'\end{CJK} means 'rate'. When come into this, the paper proposed a solution here. The work is to convert these tokens into word embeddings vector and then sum them up. Lastly, the average of embeddings is the whole representation of the multi-token input pieces. If doing so, the information could be captured entirely and is assigned to equal weight.

\subsection{Information filter}
\label{sec:if}
Based on hypothesis discussed before \cite{information}, the probability may not convey the human-like information properly. So the information theory-related strategy is reasonable to be apply in decoding strategy here. We present an information filter based on information theory, used to filter tokens that do not satisfied the required entropy out in order to enhance faithfulness. The tokens left need to harbor an information amount close to an entropy introduced below in order to convey human-like message.

Given the true probability distribution $\boldsymbol{p}$ of natural language strings, we can compute the information content of a string precisely. Assuming that $\boldsymbol{q}$ is the probability distribution predicted by the model, which approximates $\boldsymbol{p}$ well, we can use it to estimate the information content of the sentence. The entropy, i,e, information amount, of a text string $\boldsymbol{y}$ is defined as:

\begin{equation}
I(\boldsymbol{y})=-\log q(\boldsymbol{y}), \label{eq5}
\end{equation}

where $I(\boldsymbol{y})$ refers to the amount of information of $\boldsymbol{y}$, $q(\boldsymbol{y})$ is the distribution predicted to fit the true probability distribution of $\boldsymbol{y}$. Subsequently, we denote the expected information amount of a random $\boldsymbol{y}$ extracted from $\boldsymbol{q}$, also known as the entropy of $\boldsymbol{q}$ as:

\begin{equation}
\mathbb{E}_{q} [I(\boldsymbol{y})]=Ent(q)=-\sum_{\boldsymbol{y}}q(\boldsymbol{y})\log q(\boldsymbol{y}), \label{eq6}
\end{equation}

which could be perceived as the entropy of distribution obtained in each decoding step. After getting the entropy, what is the principle of selecting 'human-like token'? We suppose the information content of next token to be selected need to be close to the entropy of distribution received at certain step, i.e., the entropy given previous context, in order to convey human-like information \cite{speaking-human}. In other word, the amount of information contained in each token in sentence plays a decisive role in whether the sentence is human-like. We denote the difference, $\boldsymbol{\varepsilon}$, as:

\begin{equation}
\boldsymbol{\varepsilon_i}=|Ent(q)-I(y_i)|, \label{eq7}
\end{equation}

where $I(y_i)$ is the information amount contained in $y_i$. The token selected to be next token of sentence at each decoding step must satisfy:

\begin{equation}
\boldsymbol{\varepsilon_i} \in [Ent(q)-\boldsymbol{\varepsilon}, Ent(q)+\boldsymbol{\varepsilon}]. \label{eq8}
\end{equation}

Here, $\boldsymbol{\varepsilon}$ represents a threshold set by human indicating the maximum permitted difference. Only tokens satisfied above information amount requirement could pass the information filter, while the probabilities of tokens who dissatisfied the requirement will be set to 0 and filter out. After information filter, the probability distribution $p_{post}$ is demonstrated as:

\begin{equation}
p_{post}(i)=\begin{cases}
0, &\boldsymbol{\varepsilon_i} \notin [Ent(q)-\boldsymbol{\varepsilon}, Ent(q)+\boldsymbol{\varepsilon}]\\
\label{eq9}
p(i),&other
\end{cases} 
\end{equation}

where $p(i)$ is pre-probability before information filter. Normalization is conducted after Eq. \eqref{eq9}.

\subsection{IFDID: Information filter upon diversity-improved decoding}

\begin{figure}[ht]
\centering
\includegraphics[width=0.49\textwidth]{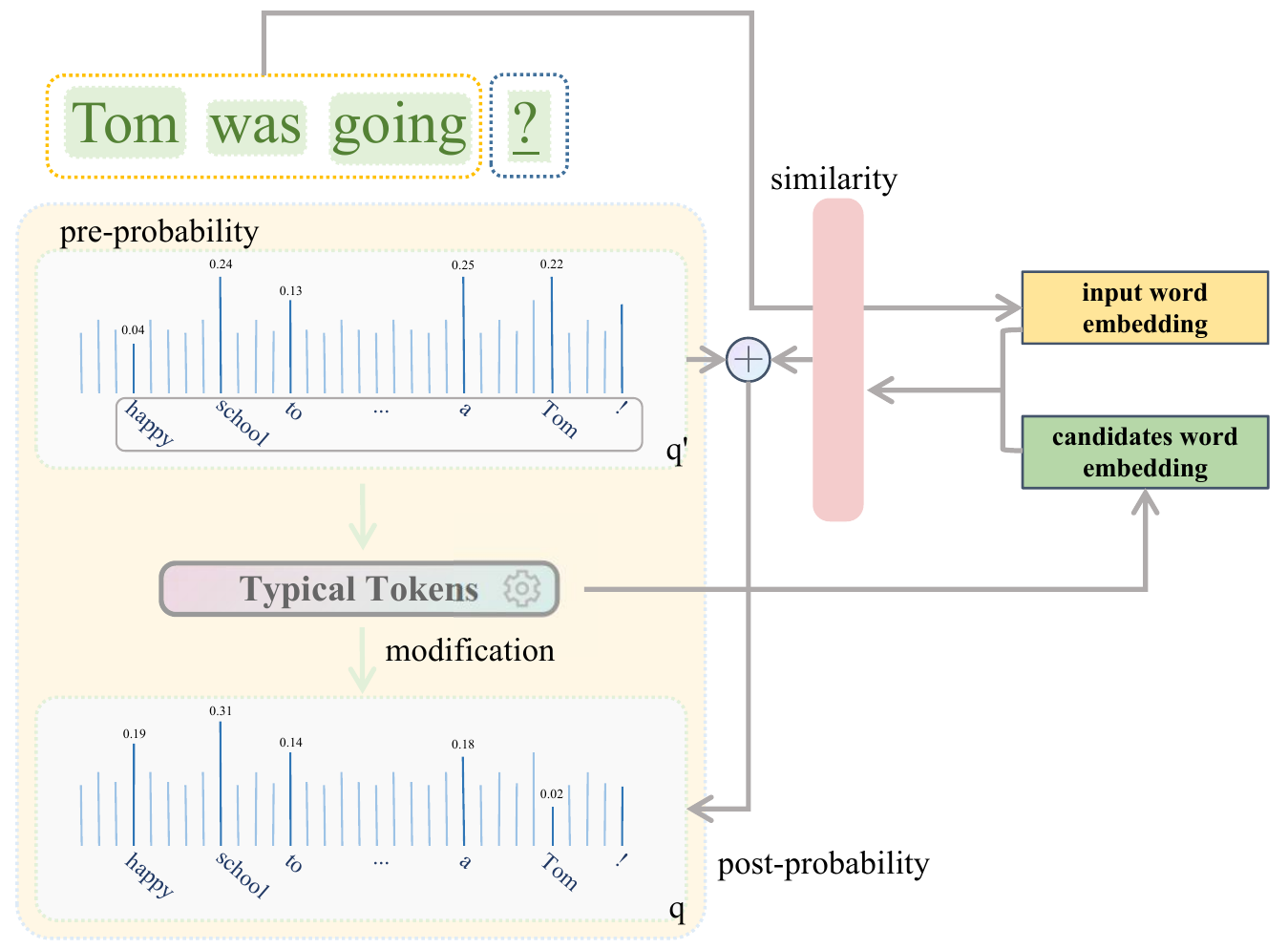}
\caption{Overview of IFDID-SIMI. It is a practical implement of Enhance stage in IFDID. The modification principle is based on word embedding similarity.}
\label{fig:fig3}
\end{figure}

After introducing proposed diversity-improved approach (DID) and information filter (IF), the overview of IFDID strategy is demonstrated in Fig. \ref{fig:fig2}, which is a two-stage decoding strategy. We demonstrate IFDID under the condition of story generation task \cite{story}. The left part is Enhance stage, a.k.a diversity-improved decoding (DID) introduced in Sec. \ref{sec:did}, which is aimed at enhance diversity through the selection and modification manipulated upon typical tokens set. The typical set includes 'happy', 'school', etc. and their probability is modified. Additionally, the right part is an information filter (IF) in Sec. \ref{sec:if} known as Filter stage committing to guarantee faithfulness. 'IM' refers to information content. IM of token 'a' does not satisfied the Eq. \eqref{eq8}, hence, it will not be selected. After IF stage, we randomly select tokens from the left tokens set, whose probability is not 0. Ultimately, 'school' is chosen to be the next token in this example.

\subsection{IFDID-SIMI: Similarity-based decoding infused with IFDID}
This section introduce a novel map function between pre-probability and post-probability in Enhance stage presented in the paper. Shown in Fig. \ref{fig:fig3}, instead of input pieces $\boldsymbol{I}$ denoted in Sec. \ref{sec:did}, a new fashion of typical tokens selection is proposed, whilst the theme tokens are designed to be a combination of {\bf near-synonyms} of input pieces. We compute the word embeddings of each tokens in input pieces $\boldsymbol{I}$ leveraging the embedding matrix taken from embedding layer and sum them up to form an average word embedding, $emb(I)$, representing input data. Subsequently, computing the similarity between input data embedding and embeddings of each token in theme tokens set, $emb(i)$, after calculating a similarity vector, based on this, we modify the pre-probability by Eq. \eqref{eq10}.

\begin{equation}
\label{eq10}
prob_i=prob_i+\lambda\cdot cos\_similarity(emb(I),emb(i)).
\end{equation}

Here, $prob_i$ refers to $\boldsymbol{i}$-th token in typical tokens set, and $\lambda$ is a hyperparameter to control the probability increment. This modification is a practical implement of Enhance stage, which could increase the likelihood of additional tokens (e.g., near-synonyms), thereby broadening the selection range and increasing the diversity of output text. Our approach is based on similarity of tokens, hence is called IFDID-SIMIlarity.

\begin{table*}[htb]   
\begin{center}
\caption{Result on CommonGEN. We report seven decoding strategies performance here. {\it greedy} means greedy search, and {\it beam.} is the abbreviation of beam search. Correspondingly, {\it temp.} is the abbreviation of temperature sampling, and {\it gamma.} refers to gamma sample. We demonstrate results on r1, r2, rl (ROUGE-1, ROUGE-2, ROUGE-L) and bleu2 (BLEU-2) for faithfulness, along with dist2 and uniq2 for diversity, whereas ppl (perplexity) for fluency. As for human evaluation, {\it flu.} is the abbreviation of fluency and similarly, {\it div.} refers to diversity, {\it faith.} means faithfulness and {\it ov.} is overall quality. The best results in each column are in {\bf bold}, and the second best ones are \underline{underlined}.}
\label{table:t1} 
\begin{tabular}{l|c c c c c c c|c c c c} 
\hline    \textbf{decoding} & \multicolumn{7}{c|}{\textbf{automatic evaluation}}& \multicolumn{4}{c}{\textbf{human evaluation}} \\
\textbf{strategy} & {\textbf{r1${\uparrow}$}} & {\textbf{r2${\uparrow}$}} & {\textbf{rl${\uparrow}$}} & {\textbf{bleu2${\uparrow}$}}  & {\textbf{dist2${\uparrow}$}} & {\textbf{uniq2${\uparrow}$}} & {\textbf{ppl${\downarrow}$}} & \textbf{flu.${\downarrow}$} & \textbf{div.${\downarrow}$} & \textbf{faith.${\downarrow}$} & \textbf{ov.${\downarrow}$} \\
\hline  greedy & 29.57 & \underline{9.97} & 26.82 & 21.48 & 0.39 & 3573 & 4.15 & 1.46 & 1.19 & 1.09 & 1.56\\
beam. & \underline{30.01} & \textbf{10.02} & \textbf{27.19} & \textbf{22.24} & 0.38 & 3508 & 4.27 & \textbf{1.36} & 1.15 & \textbf{1.06} & \textbf{1.45} \\
temp. & 28.81 & 9.18 & 25.88 & 20.80 & 0.44 & 4164 & 4.43 & \underline{1.44} & 1.17 & 1.09 & \underline{1.51}\\
top-$k$ & \textbf{30.05} & 9.63 & \underline{26.91} & \underline{21.87} & 0.44 & 4213 & 4.37 & 1.62 & 1.18 & 1.09 & 1.68\\
gamma. & 28.41 & 8.71 & 24.65 & 19.23 & \underline{0.53} & \underline{4975} & 4.36 & 1.71 & \textbf{1.1} & 1.1 & 1.71\\
IFDID & 29.93 & 9.02 & 25.82 &20.33 & 0.48 & 4676 & \textbf{4.10} & 1.56 & 1.16 & \underline{1.07} & 1.56\\
IFDID-SIMI & 28.86 & 8.45 & 24.78 & 19.08 & \textbf{0.54} & \textbf{5528} & \underline{4.26} & 1.67 & \underline{1.13} & 1.09 & 1.7\\
\hline
\end{tabular}   
\end{center}   
\end{table*}

\section{Experiments}
\label{sec:exp}

In this section, we explore the effectiveness towards tradeoff between diversity and faithfulness of our presented decoding strategy, IFDID, on three general natural language generation tasks: commonsense generation (Sec. \ref{ssec:commongen}), story generation (Sec. \ref{ssec:story}) and specific style text generation (Sec. \ref{ssec:adgen}). Upon all tasks above, compared with beam search, sampling strategy and guided decoding, IFDID outperforms beam search and sampling strategy upon diversity, while outperforms guided decoding on faithfulness. Alongside, we evaluate IFDID-SIMI as well, it outperforms IFDID on diversity, but sacrifice more faithfulness to strike the balance. See Sec. \ref{ssec:human} for human evaluation explanation.

\noindent{\bf Evaluation metrics.} We assess our decoding strategy on faithfulness, diversity and fluency utilizing automatic evaluation and human evaluation. We leverage following automatic evaluation metrics that substantially used to evaluate the quality of generated text to evaluate respectively upon three tasks: ROUGE \cite{ROUGE}, BLEU \cite{bleu}\footnote{We use \url{Pyrouge} package to evaluate ROUGE as well as \url{corpus\_bleu} to assess BLEU value.}, Dist \cite{dist}, Uniq (unique ngrams) and perplexity \cite{ppl}. We use ROUGE and BLEU to represent faithfulness, while Dist and Uniq refers to diversity as well as perplexity is the scale of fluency. Besides, we conduct human evaluation as well, more details could be found in Sec. \ref{ssec:human}.

\subsection{Commonsense generation}
\label{ssec:commongen}

{\bf Dataset and Baselines.} We conduct experiment upon CommonGEN \cite{commongen}, a publicly recognized constrained commonsense generation task. Given a collection of common concepts (e.g, climb, wall, talk), the aim is to produce a cohesive phrase expressing an everyday occurrence using these terms (e.g, 'The lady who is talking teaches her pupils how to climb the wall.'). We employ the T5 model \cite{t5} fine-tuned on CommonGEN\footnote{\url{https://huggingface.co/mrm8488/t5-base-finetuned-common_gen}} available on HuggingFace platform\footnote{\url{https://huggingface.co}}. Moreover, we compare with previous proposed decoding algorithms, including greedy search, beam search\footnote{Here, the beam search strategy we leverage the forbid\_duplicate\_ngrams strategy, namely, we prohibit repeated ngrams from appearing twice in a sentence.} \cite{beam}, temperature sampling, top-$k$ sampling and gamma sample \cite{gamma}.

\noindent {\bf Setup and Results.} We found $beamsize=5$ is approximate for beam search algorithm, since the resource usage is appropriate and could be distinguished from greedy search. As for temperature sampling and top-$k$ sampling, we conduct a series attempts to find the best parameters and found $t$=0.5, $k$=5 respectively could provide with best result. For gamma sample, due to the number of parameters selection is vast, we follow \cite{gamma} and set parameter list as: $\gamma_{rep}$=0.99, $\gamma_{topic}$=0.4, $\gamma_{sentence}$=0.9 and top-$n$=350. We set IFDID's $\varepsilon$ to 0.1 and IFDID-SIMI's $\lambda$ to 0.0005.

The result is demonstrated in Table \ref{table:t1}, including result of automatic evaluation and human evaluation respectively. The vital observation is as follows. More generated cases could be found in Appendix \ref{sec:app1}. As for details about human evaluation, see Sec. \ref{ssec:human}.

\begin{figure*}[ht]
\centering
\begin{minipage}[b]{0.32\linewidth}
  \centering
  \centerline{\includegraphics[width=5.63cm]{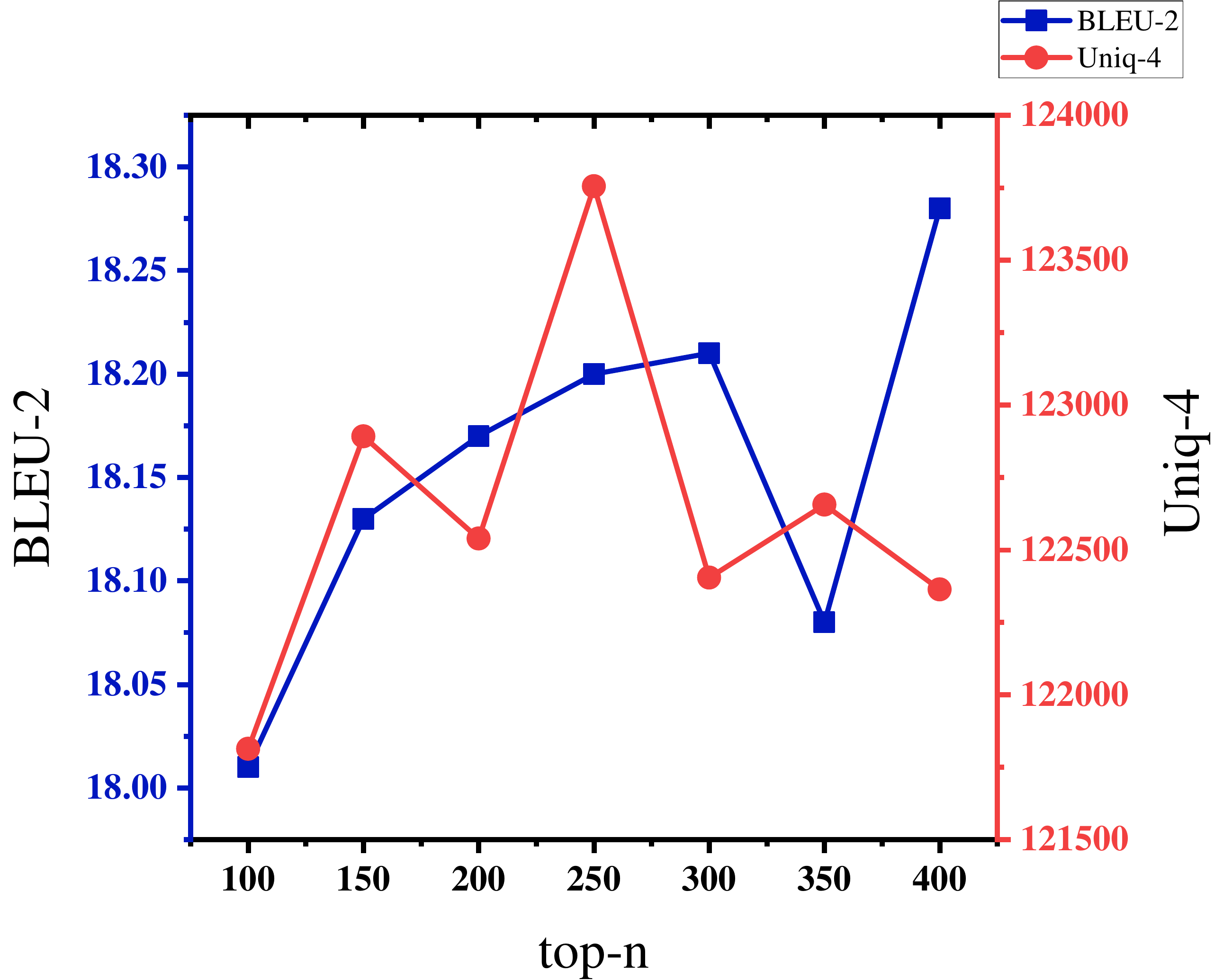}}
  \centerline{(a)}\medskip
\end{minipage}
\hfill
\begin{minipage}[b]{0.32\linewidth}
  \centering
  \centerline{\includegraphics[width=5.63cm]{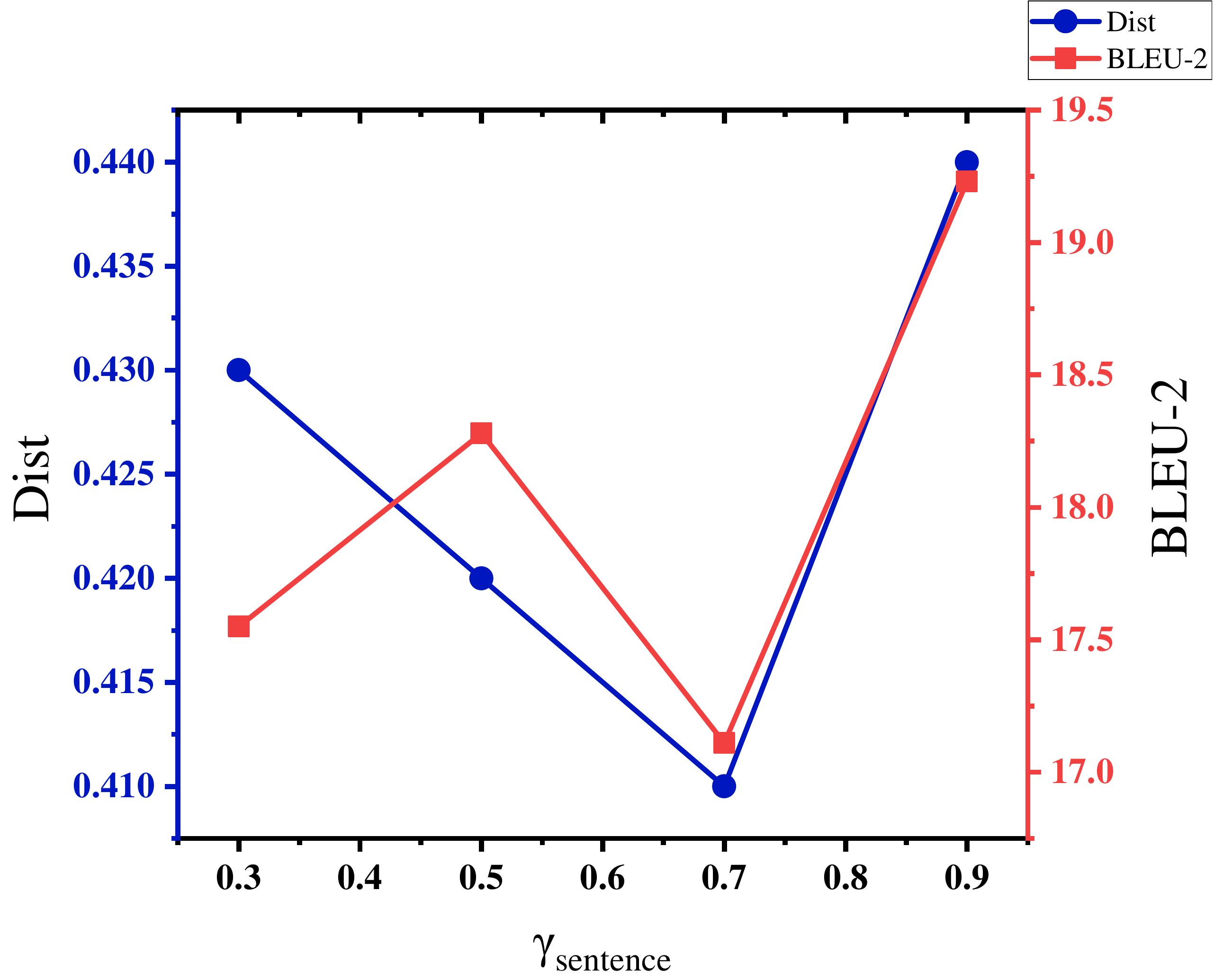}}
  \centerline{(b)}\medskip
\end{minipage}
\begin{minipage}[b]{0.34\linewidth}
  \centering
  \centerline{\includegraphics[width=5.63cm]{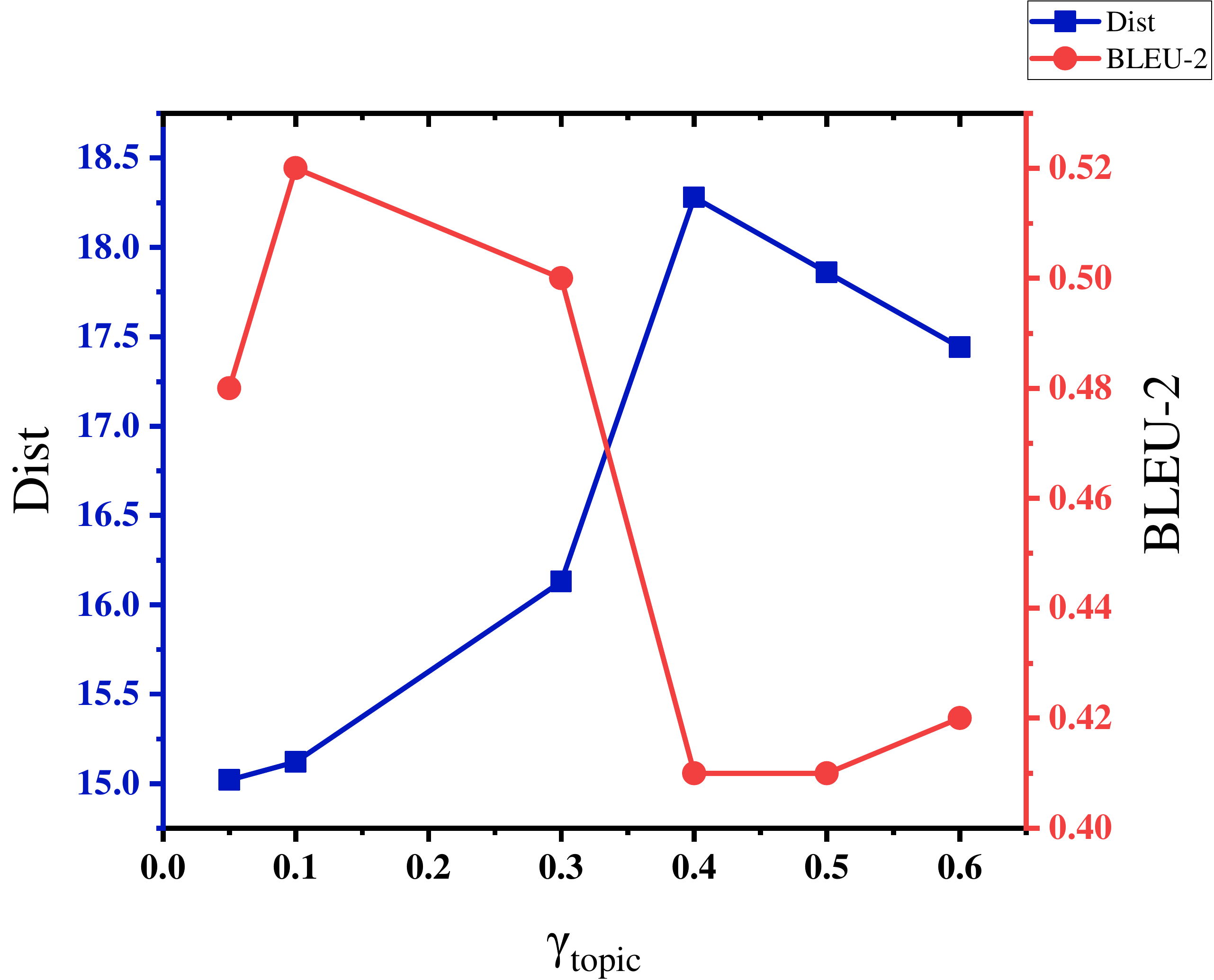}}
  \centerline{(c)}\medskip
\end{minipage}
\caption{The performance of IFDID or IFDID-SIMI in terms of diversity (represented by Dist and Uniq) and faithfulness (represented by BLEU) under different parameter settings. X-axis is the controlled parameters, whereas y-axis is the evaluation metrics.}
\label{fig:fig4}
\end{figure*}

{\bf IFDID provides a better tradeoff between diversity and faithfulness than traditional approaches and guided decoding.} Beam search and greedy search outperform other decoding strategies in terms of faithfulness, and they generally perform best in ROUGE and BLEU metrics, followed by sampling approaches, while guided decoding such as gamma sample performs poorly in faithfulness. As for IFDID, our proposed decoding strategy, performs between sampling approaches and guided decoding in terms of faithfulness. Furthermore, at the diversity level, the worst performers are beam search and greedy search, which perform badly in Dist and Uniq metrics because they tend to select only tokens with high probability, resulting in dull and repetitive selection results; sampling approaches also perform poorly, but outperform beam search. The best performer is guided decoding, but the increase in diversity comes at the expense of faithfulness, whereas IFDID performs significantly better in terms of diversity than beam search and greedy search, as well as sampling approaches, which, like faithfulness, performs between sampling approaches and guided decoding. Considering the two dimensions of faithfulness and diversity, beam search, greedy search and sampling approaches always produce more conservative sentences, whilst sampling approaches are less conservative than beam search and greedy search. On the contrary, guided decoding tends to generate more diversified and linguistically rich sentences, but performs poorly in the degree of input sentence coverage, i.e., faithfulness, and tends to generate unfaithful content. Our proposed IFDID, on the other hand, can balance these two metrics and significantly enhance the diversity compared with the traditional method, while compensating for the lack of faithfulness of guided decoding, and is the decoding strategy with the best tradeoff between diversity and faithfulness. Ultimately, we measured the coherence of the generated text by the perplexity metric, and beam search and greedy search outperformed sampling approaches and guided decoding to generate relatively smooth sentences, while IFDID outperformed all other decoding strategies to generate substantially smooth and readable results.

{\bf IFDID-SIMI may be viewed as an additional method of weighing diversity and faithfulness.} The faithfulness of IFDID-SIMI is comparable to that of guided decoding, but it can boost diversity somewhat, and can generate more diverse text than guided decoding with the same level of faithfulness. Compared with traditional approaches, we can conclude similarly to IFDID: IFDID-SIMI improves diversity but reduces faithfulness to a small extent.

\noindent {\bf Further Analysis.} To further analyze the effects of each parameter in IFDID on diversity and faithfulness, we separately explored $\gamma_{sentence}$, representing the parameter leveraged to control terminal tokens set, $\gamma_{topic}$, the parameter controlling theme tokens, and top-$n$, which used in IFDID-SIMI to determine the selection range of near-synonym upon CommonGEN \cite{commongen} task. The results is illustrated in Fig \ref{fig:fig4}. We can see that under the $\gamma_{sentence}$=0.9 condition, both diversity and faithfulness reach the optimal result, while both diversity and faithfulness are impaired when the generated text is shorter, namely, when the $\gamma_{sentence}$ value is smaller. The selection of $\gamma_{topic}$ plays a decisive role in the tradeoff between diversity and faithfulness. When the $\gamma_{topic}$ value is small, diversity is poor, while conversely, when the $\gamma_{topic}$ value is large, faithfulness decreases significantly. Ultimately, it is important to choose a suitable top-$n$ value to ensure the faithfulness, and basically, the larger the top-$n$, the higher the diversity.

\subsection{Story generation}
\label{ssec:story}

\begin{table*}[htb]   
\begin{center}   
\caption{Performance of different decoding strategies on RocStories test set. We report seven decoding strategies performance here. {\it nucleus.} denotes the abbreviation of nucleus sampling. As for human evaluation, {\it gra.} refers to evaluation on grammatically correctness, whereas {\it int.} is interestingness. The best results in each column are in {\bf bold}, and the second best ones are \underline{underlined}.}
\label{table:t2} 
\begin{tabular}{l|c c c c c c c|c c c c c} 
\hline    \textbf{decoding} & \multicolumn{7}{c|}{\textbf{automatic evaluation}}& \multicolumn{5}{c}{\textbf{human evaluation}} \\
\textbf{strategy} & {\textbf{r1${\uparrow}$}} & {\textbf{r2${\uparrow}$}} & {\textbf{rl${\uparrow}$}} & {\textbf{bleu2${\uparrow}$}}  & {\textbf{dist2${\uparrow}$}} & {\textbf{uniq2${\uparrow}$}} & {\textbf{ppl${\downarrow}$}} & {\textbf{gra.${\uparrow}$}} & {\textbf{flu.${\uparrow}$}} & {\textbf{faith.${\downarrow}$}} & \textbf{int.${\downarrow}$} & \textbf{ov.${\downarrow}$}\\
\hline  greedy & 17.78 & 2.80 & 13.44 & 11.35 & 0.14 & 10165 & 46.01 & 1.07 & 1.77 & 1.62 & 3.32 & 2.19\\
top-$k$ & 23.25 & \textbf{3.53} & \textbf{16.39} & 10.03 & 0.16 & 16322 & 41.54 & \underline{1.04} & 1.9 & \textbf{1.47} & 3.02 & 2.13\\
temp. & 20.88 & 3.25 & 15.17 & \underline{11.74} & 0.17 & 13930 & \underline{40.43} & 1.08 & 1.78 & \underline{1.49} & 3.1 & 2.13\\
nucleus. & 20.83 & \underline{3.42} & 15.35 & \textbf{12.09} & 0.16 & 12650 & 42.22 & \textbf{1.01} & 1.81 & 1.59 & 2.99 & 2.12\\
gamma. & 24.37 & 2.39 & 14.78 & 8.49 & \underline{0.28} & \underline{33339} & 41.52 & \underline{1.04} & \underline{1.35} & 1.54 & \underline{2.56} & \underline{1.8}\\
IFDID & \underline{25.61} & 2.64 & \underline{15.46} & 9.04 & 0.26 & 30729 & \textbf{39.54} & \textbf{1.01} & \textbf{1.31} & 1.52 & \textbf{2.40} & \textbf{1.72}\\
IFDID-SIMI & \textbf{26.17} & 1.62 & 14.12 & 6.57 & \textbf{0.43} & \textbf{90943} & 40.89 & 1.49 & 1.82 & 1.75 & 2.83 & 2.17\\
\hline
\end{tabular} 
\end{center}   
\end{table*}

\noindent {\bf Dataset and Baselines.} Our experiment is carried upon RocStories\footnote{The data could be found at \url{https://github.com/yxuansu/SimCTG/tree/main/data/ROCStories}} \cite{rocstories} to test the ability of story generation. The dataset is annotated into specific form containing 1.5K samples, while the task is to generate stories based on the given keywords. We reproduce the available fine-tuned GPT-2 \cite{gpt2} generation model\footnote{\url{https://huggingface.co/cambridgeltl/simctg_rocstories}} on the Huggingface platform. In order to conduct adequate experiments, we compare our approaches with greedy search, top-$k$ sampling, nucleus sampling \cite{topp} as well as gamma sample \cite{gamma} to prove the performance.

\noindent {\bf Setup and Results.} Preliminarily, we set $k$=5 for top-$k$ sampling and $p$=0.3 for nucleus sampling. As for gamma sample, we conduct a series of experiments to find optimal parameters, which are as follows: $\gamma_{rep}$=0.99, $\gamma_{topic}$=0.4, $\gamma_{sentence}$=0.7 and top-$n$=300. The $\varepsilon$ for information filter in IFDID is 0.2 as well as IFDID-SIMI's $\lambda$ is 0.0005.

The result of our experiments is shown in Table \ref{table:t2}. The key observation is listed below and more analysis about human evaluation could be found in Sec. \ref{ssec:human}, whereas more cases is demonstrated in Appendix \ref{sec:app2}. From the data in the table, it can be seen that for the task of story generation, the most conservative decoding strategy of choice, greedy search, performs poorly in both diversity and faithfulness. In contrast, sampling approaches perform better in terms of faithfulness, especially nucleus sampling, and outperforms all seven decoding strategies in the BLEU metric. Guided decoding, on the other hand, while significantly outperforming traditional strategies in terms of diversity, has a very significant drop in performance in terms of faithfulness, especially in the BLEU metric. Our proposed method, IFDID, performs well in ROUGE scores, with ROUGE-L second only to top-k sampling, and has a significant improvement in faithfulness compared to the guided decoding strategy, while diversity only decreases by 0.02, which still has a large lead over traditional strategies. The IFDID-SIMI strategy, however, provides a more diversified output text at a same cost of faithfulness as guided decoding. The results confirm that IFDID and IFDID-SIMI could obtain a better tradeoff between diversity and faithfulness.

\noindent {\bf Case Study.} We present generated samples of IFDID, IFDID-SIMI and other decoding strategies given a input topic in Table \ref{table:app2}. From the results, we can see that the length of each clause in the story paragraph generated by nucleus sampling is very short, containing relatively little information, and the grammar is very simple, as well as the sentence structure is very single, albeit it is very relevant, but the last two sentences are completely consistent except for the subject. Similarly, the same problem is found in the case of the greedy search and temperature sampling, where the generated stories are very boring and uninteresting, and the length is relatively short. While top-$k$ sampling generates longer paragraphs, repetitive content such as "was glad he skipped school and got into better shape" seriously affects the quality of the generated stories. On the contrary, the stories generated by the guided decoding strategy such as gamma sample have higher content fluency, but the content deviates from the given topic. In contrast, IFDID, our proposed method, is a story of high overall quality, both in terms of story fluency and interestingness, with a twist plot, moderate length, and close to the topic. The content generated by the IFDID-SIMI strategy is richer and the story is intriguing and spine-tingling, however, everything is fine except for one small defect of having capitalization in the middle of a sentence.

\subsection{Specific style text generation}
\label{ssec:adgen}

\begin{table*}[htb]   
\begin{center}   
\caption{Diversity, faithfulness, fluency and overall quality performance on AdGen test set. We report the performance of seven decoding strategies. $beam_{without}$ means beam search without the forbid\_duplicate\_ngrams strategy, which is leveraged to reduce repetition. The best results in each column are in {\bf bold}, and the second best ones are \underline{underlined}.}
\label{table:t3} 
\begin{tabular}{l|c c c c c c c|c c c c} 
\hline    \textbf{decoding} & \multicolumn{7}{c|}{\textbf{automatic evaluation}}& \multicolumn{4}{c}{\textbf{human evaluation}} \\
\textbf{strategy} & {\textbf{r1${\uparrow}$}} & {\textbf{r2${\uparrow}$}} & {\textbf{rl${\uparrow}$}} & {\textbf{bleu2${\uparrow}$}}  & {\textbf{dist4${\uparrow}$}} & {\textbf{uniq4${\uparrow}$}} & {\textbf{ppl${\downarrow}$}} & {\textbf{flu.${\uparrow}$}} & {\textbf{faith.${\downarrow}$}} & \textbf{div.${\downarrow}$} & \textbf{ov.${\downarrow}$}\\
\hline  beam${_{without}}$ & 23.49 & 7.66 & 18.73 & 9.43 & 0.02 & 8074 & 4.38 & 1.03 & 2.98 & 3 & 2.98\\
beam. & 31.26 & \textbf{12.40} & 19.34 & 17.06 & 0.08 & 17079 & \underline{1.49} & \underline{2.27} & \textbf{1.94} & 1.94 & \underline{1.99}\\
top-$k$ & 32.92 & 11.16 & \underline{20.37} & 18.22 & 0.24 & 64339 & 1.57 & 1.98 & \underline{2.14} & 2.08 & 2.05\\
nucleus. & 33.30 & \underline{11.26} & \textbf{20.45} & \underline{18.39} & 0.25 & 64661 & 1.54 & 2.03 & 2.15 & 2.08 & 2.09\\
gamma. & 34.37 & 10.28 & 19.04 & 18.28 & \underline{0.41} & \underline{122363} & \textbf{1.43} & 2.2 & 2.72 & \underline{1.19} & 2.04\\
IFDID & \textbf{35.37} & 11.09 & 19.26 & \textbf{18.83} & 0.35 & 96238 & \textbf{1.43} & \textbf{2.28} & 2.51 & 1.29 & \textbf{1.98}\\
IFDID-SIMI & \underline{34.48} & 10.22 & 18.98 & 18.32 & \textbf{0.45} & \textbf{138953} & 1.61 & 1.9 & 2.85 &\textbf{1.08} & 2.25\\
\hline
\end{tabular}   
\end{center}   
\end{table*}

\noindent {\bf Dataset and Baselines.} We conduct the specific style text generation experiment on AdGen \cite{adgen}, a Chinese advertising text generation constructed from a Chinese e-commerce platform. The dataset contains 119K advertisement text and apparel specification table pairings. Each table has a collection of attribute-value pairs that describe an item of apparel. The task is to generate text with advertising style based on the given descriptive information. We use 3127 samples for testing and leverage the OPPO\footnote{\url{https://www.oppo.com/}} self-research model, which is not open-sourced. The baselines include beam search \cite{beam} with and without the forbid\_duplicate\_ngrams strategy, top-$k$ sampling, nucleus sampling \cite{topp} and gamma sample \cite{gamma}.

\noindent {\bf Setup and Results.} We set the $beamsize=5$, which is appropriate for this task, while $k$=5 for top-$k$ sampling and $p$=0.3 for nucleus sampling. Moreover, we set a parameter list for gamma sample: $\gamma_{rep}$=0.9, $\gamma_{topic}$=0.4, $\gamma_{sentence}$=0.5 and top-$n$=300. Ultimately, we set $\varepsilon$=0.95 for IFDID as well as $\lambda$=0.001 for IFDID-SIMI.

The result is demonstrated in Table \ref{table:t3}. Firstly, on the specific style text generation task, the beam search without forbid\_duplicate\_ngrams strategy produces substantially poor results on diversity and faithfulness, and the generated text is disfluent with very high perplexity values. This is due to the degeneration phenomenon caused by the beam search approach. Once adding the forbid\_duplicate\_ngrams strategy, the faithfulness is greatly improved, but the diversity still performs very poorly. Various Sampling approaches show comparable effects on diversity, and the faithfulness is on par with beam search. Guided decoding, a.k.a gamma sample, performs worse than traditional strategies in terms of faithfulness, but the diversity has been greatly improved. Our proposed IFDID compensates the sacrifice of gamma sample on faithfulness, outperforming all other decoding strategies in BLEU metrics, and the fluency also beats other decoding strategies. The IFDID-SIMI strategy, on the other hand, exhibits the highest diversity, and the faithfulness is at the same level as gamma sample. The results on this experiment demonstrate that our proposed IFDID and IFDID-SIMI decoding strategies can successfully balance diversity and faithfulness on both English and Chinese datasets.

\subsection{Human evaluation}
\label{ssec:human}

This section give a detailed explanation of human evaluation results\footnote{We give more details in Appendix \ref{sec:app-human}.}. We recruit a group of people to be reviewer and we first determined the samples needed for the human evaluation: samples generated by 7 decoding strategies for each of the three tasks, which contains a total of 21 groups of samples, and 2100 samples in total. Then, we randomly shuffle these samples to construct 21 manual evaluation sheets, each containing 100 randomly scrambled samples. The principle is that each reviewer receives a human evaluation sheet containing only ONE task, but the sheets contain text generated by different decoding strategies, and the reviewer scores the samples on the following four dimensions without knowing in advance which decoding strategy each sample comes from:

\noindent {\bf Fluency.} Fluency is a metric directly linking with readability, which is a key component of human-like text. We collectively evaluate fluency from semantic and syntactic aspects.

\noindent {\bf Diversity.} Diversity refers to whether a sentence contains a variety of words, sentence forms and expressions without being dull and repetitive.

\noindent {\bf Faithfulness.} Faithfulness refers to whether the generated text matches the input text. For CommonGEN and AdGen task, faithfulness refers to whether the generated hypothesis contains all the words in the input data pieces; for RocStories, it refers to whether the generated story is relevant to the topic.

\noindent {\bf Grammar.} It refers to whether the generated sentence is grammatically correct.

\noindent {\bf Overall quality.} It is the score given by the reviewer for the overall quality of the sentence generation based on the above three metrics.

Reviewers scored these dimensions on a scale from 1 to 3, and for interestingness, scores are from 1 to 5. The meaning of score could be seen in Appendix \ref{sec:app-human}.

\section{Further discussion about degeneration in NLG}
\label{sec:degeneration}

This section provide more discussion about degeneration occurring in NLG tasks. Upon the AdGen task (Sec. \ref{ssec:adgen}), we could clearly observe in Table \ref{table:app4} that while we apply beam search without reducing repetition, the degeneration occurs. Additionally, the experiment upon RocStories (Sec. \ref{ssec:story}) shows that some repetitive expressions appear when using top-$k$ sampling (see in Table \ref{table:app2}). In order to explore the condition of degeneration phenomenon, we conduct an extra experiment demonstrated here.

\begin{figure}[ht]
\centering
\includegraphics[width=0.49\textwidth]{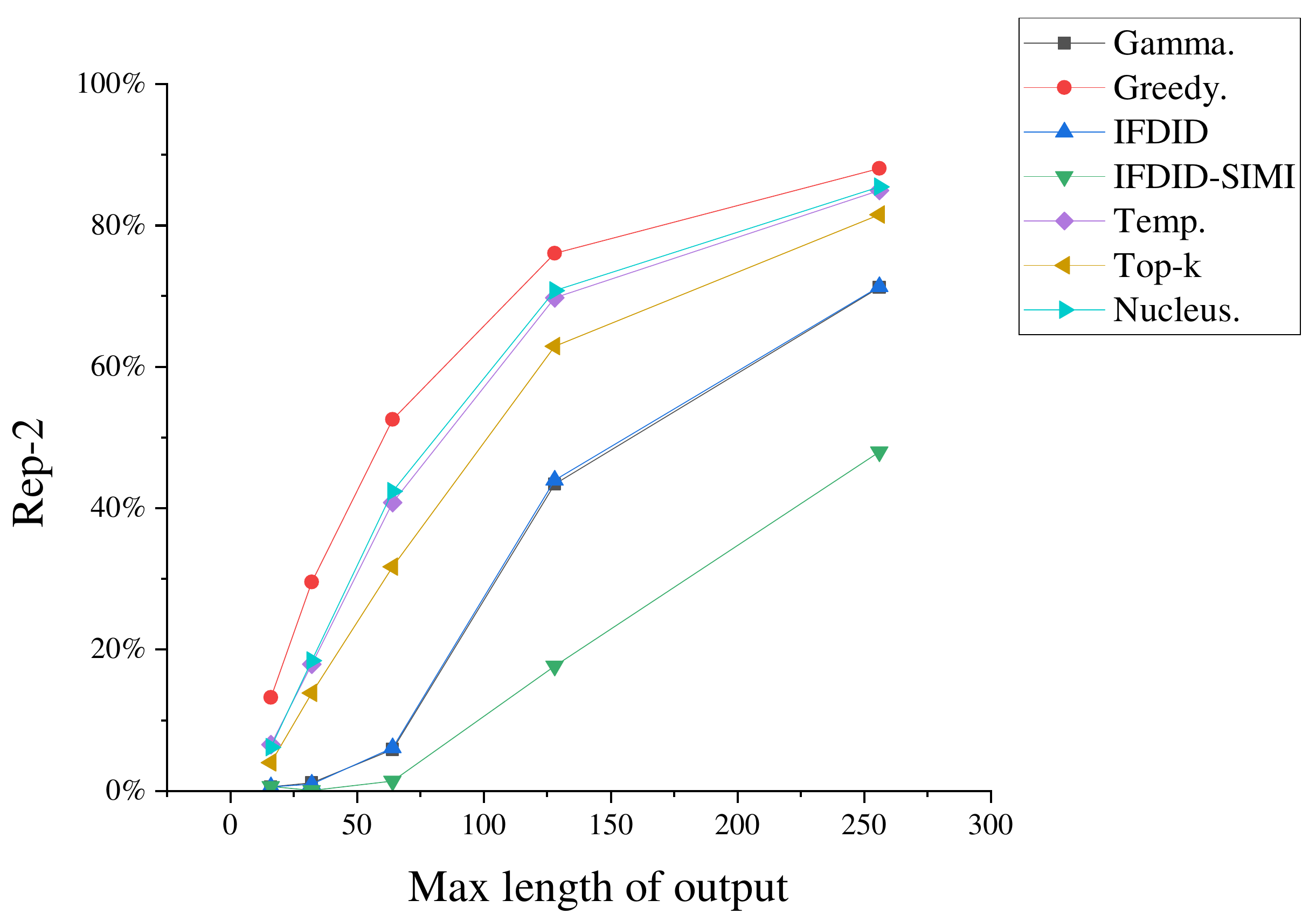}
\caption{The Rep-2 metrics (the statistic of duplicate 2-gram) of different decoding strategy under the condition of various max sequence length. The text that exceeds the maximum distance is truncated.}
\label{fig:fig5}
\end{figure}

On the rocstories dataset, we analyzed 1500 stories generated by each of the seven decoding strategies. We control the maximum length of the generated text when the model generates sentences, and any text exceeding this limit will be truncated. The results are shown in Fig. \ref{fig:fig5}, where we can see that the repetition rate of the greedy search strategy is the largest for each maximum length limit, followed by sampling approaches. Besides, our proposed IFDID and gamma sample show almost the same repetition rate, and IFDID-SIMI gets the lowest repetition rate here. However, it is noteworthy that in terms of the gradient of each line, the traditional decoding strategies reach their maximum gradient as early as the length interval of 10-30, while the IFDID and IFDID-SIMI decoding strategies gradient reach their maximum value only when the maximum generated sequence length is limited to 60-130. From the case study in Table \ref{table:app2}, it can be seen that when the gradient of each line in Fig. \ref{fig:fig5} is maximum, it indicates the degeneration phenomenon, i.e., the generated text starts to repeat completely. However, the moment of degeneration occurring is different for each decoding strategy, which is an interesting phenomenon and illustrates that degeneration is closely related to the length of the generated sequence and its limitation. Our supplementary experiments provide preliminary possibilities and supporting evidence to further investigate the factors influencing degeneration.

\section{Conclusion}
\label{sec:conclusion}
In this paper, we propose a decoding strategy, IFDID, in order to obtain a tradeoff between diversity and faithfulness. Our approach is a two-stage decoding strategy leveraging Enhance-Filter framework, which can balance faithfulness and diversity at the mean time. Our experiments cover a wide range of tasks, illustrating that it is currently the SOTA decoding strategy for pursuing tradeoff between diversity and faithfulness. We also present IFDID-SIMI, another decoding strategy to promote diversity with less faithfulness loss, which could be perceived as another practical implementation of Enhance-Filter framework. Besides, we give a theoretical analysis of degeneration and empirically explore the relation with output sequence length. 

\noindent {\bf Future Works.} In the future, our study will focus on finding more application scenarios for this decoding strategy, such as summarization, automatic dialogue and other practical scenarios. Additionally, we will continue to investigate other factors related to the degeneration phenomenon.

\section{Acknowledgements}
\label{sec:ack}
We'd like to thank reviewers for their valuable feedback. We specially thank Tensor Lab of OPPO Research Institute for support with technical discussion. Many thanks also to people in OPPO Research Institute who help us for human evaluation.

\section{Ethical consideration}
\label{sec:eth}

When exploring applications, a variety of ethical issues arise due to the nature of text generation. The fundamental cause of failure is that the model might replicate undesirable target qualities in the training data.

\noindent {\bf Deviations in data and faithfulness consideration.} For the majority of NLP jobs, biases in the data related to gender, ethnicity, etc. risk being propagated in the model predictions. This is particularly relevant if the models were developed using outdated data that does not accurately reflect modern attitudes and behaviors. Secondly, as for decoding strategy, the selection of near-synonym words can substantially affect the quality of the generated text of the model. If the near-synonym words are not selected properly, the generated content will also become untrustworthy.

The model is only learning to act like its underlying source material, hence the aforementioned arguments are not malicious. Furthermore, generation models may be abused for bad purposes. These include creating spam, fake news, and other texts intended to deceive the majority of the population.

\vfill\pagebreak

\bibliographystyle{IEEEbib}
\bibliography{strings}

\begin{thebibliography}{10}

\bibitem{commongen}
B.~Y. Lin, W.~Zhou, M.~Shen, P.~Zhou, C.~Bhagavatula, Y.~Choi, and X.~Ren,
\newblock ``{C}ommon{G}en: A constrained text generation challenge for generative commonsense reasoning,''
\newblock in {\em Findings of the Association for Computational Linguistics: EMNLP 2020}, Online, Nov. 2020, pp. 1823--1840, Association for Computational Linguistics.

\bibitem{nlgsurvey}
A.~Gatt and E.~Krahmer,
\newblock ``Survey of the state of the art in natural language generation: Core tasks, applications and evaluation,''
\newblock {\em Journal of Artificial Intelligence Research}, vol. 61, pp. 65--170, January 2018.

\bibitem{sum}
I.~Mani,
\newblock {\em Automatic Summarization}, vol.~3,
\newblock John Benjamins Publishing, 2001.

\bibitem{sum2}
J.~M. Torres-Moreno,
\newblock {\em Automatic Text Summarization},
\newblock John Wiley \& Sons, October 2014.

\bibitem{Hutchins1992AnIT}
W.~J. Hutchins and H.~L. Somers,
\newblock ``An introduction to machine translation,''
\newblock 1992.

\bibitem{zhou2017neural}
Q.~Zhou, N.~Yang, F.~Wei, C.~Tan, H.~Bao, and M.~Zhou,
\newblock ``Neural question generation from text: A preliminary study,''
\newblock in {\em National CCF Conference on Natural Language Processing and Chinese Computing}. Springer, 2017, pp. 662--671.

\bibitem{evaluation}
A.~Celikyilmaz, E.~Clark, and J.~Gao,
\newblock ``Evaluation of text generation: A survey,''
\newblock {\em Arxiv preprint}, vol. abs/2006.14799, 2018.

\bibitem{composition}
S.~Narayan, G.~Sim{\~o}es, Y.~Zhao, J.~Maynez, D.~Das, M.~Collins, and M.~Lapata,
\newblock ``A well-composed text is half done! composition sampling for diverse conditional generation,''
\newblock in {\em Proceedings of the 60th Annual Meeting of the Association for Computational Linguistics (Volume 1: Long Papers)}, Dublin, Ireland, May 2022, pp. 1319--1339, Association for Computational Linguistics.

\bibitem{bleu}
K.~Papineni, S.~Roukos, T.~Ward, and W.~Zhu,
\newblock ``Bleu: a method for automatic evaluation of machine translation,''
\newblock in {\em Proceedings of the 40th Annual Meeting of the Association for Computational Linguistics (ACL)}, Philadelphia, July 2002, pp. 311--318, Association for Computational Linguistics.

\bibitem{story}
A.~Fan, M.~Lewis, and Y.~Dauphin,
\newblock ``Hierarchical neural story generation,''
\newblock in {\em Proceedings of the 56th Annual Meeting of the Association for Computational Linguistics (Volume 1: Long Papers)}, Melbourne, Australia, July 2018, pp. 889--898, Association for Computational Linguistics.

\bibitem{astar}
X.~Lu, S.~Welleck, P.~West, L.~Jiang, J.~Kasai, D.~Khashabi, R.~L. Bras, L.~Qin, Y.~Yu, R.~Zellers, N.~A. Smith, and Y.~Choi,
\newblock ``{N}euro{L}ogic a*esque decoding: Constrained text generation with lookahead heuristics,''
\newblock in {\em Proceedings of the 2022 Conference of the North American Chapter of the Association for Computational Linguistics: Human Language Technologies}, Seattle, United States, July 2022, pp. 780--799, Association for Computational Linguistics.

\bibitem{data2text}
T.~C. Ferreira, C.~Lee, E.~Miltenburg, and E.~Krahmer,
\newblock ``Neural data-to-text generation: A comparison between pipeline and end-to-end architectures,''
\newblock in {\em Proceedings of the 2019 Conference on Empirical Methods in Natural Language Processing and the 9th International Joint Conference on Natural Language Processing (EMNLP-IJCNLP)}, Hong Kong, China, Nov. 2019, pp. 552--562, Association for Computational Linguistics.

\bibitem{paraphrase}
N.~Madnani and B.~J. Dorr,
\newblock ``{Generating Phrasal and Sentential Paraphrases: A Survey of Data-Driven Methods},''
\newblock {\em Computational Linguistics}, vol. 36, no. 3, pp. 341--387, 09 2010.

\bibitem{eCommerce}
Q.~Chen, J.~Lin, Y.~Zhang, H.~Yang, J.~Zhou, and J.~Tang,
\newblock ``Towards knowledge-based personalized product description generation in e-commerce,''
\newblock in {\em Proceedings of the 25th ACM SIGKDD International Conference on Knowledge Discovery \& Data Mining}, New York, NY, USA, 2019, KDD '19, p. 3040–3050, Association for Computing Machinery.

\bibitem{deduplication}
B.~Gyawali, L.~Anastasiou, and P.~Knoth,
\newblock ``Deduplication of scholarly documents using locality sensitive hashing and word embeddings,''
\newblock in {\em Proceedings of the 12th Language Resources and Evaluation Conference}, Marseille, France, May 2020, pp. 901--910, European Language Resources Association.

\bibitem{beam}
J.~Li, W.~Monroe, A.~Ritter, D.~Jurafsky, M.~Galley, and J.~Gao,
\newblock ``Deep reinforcement learning for dialogue generation,''
\newblock in {\em Proceedings of the 2016 Conference on Empirical Methods in Natural Language Processing}, Austin, Texas, Nov. 2016, pp. 1192--1202, Association for Computational Linguistics.

\bibitem{topp}
A.~Holtzman, J.~Buys, L.~Du, M.~Forbes, and Y.~Choi,
\newblock ``The curious case of neural text degeneration,''
\newblock in {\em 8th International Conference on Learning Representations, {ICLR} 2020, Addis Ababa, Ethiopia, April 26-30, 2020}. 2020, OpenReview.net.

\bibitem{unlikelihood}
S.~Welleck, I.~Kulikov, S.~Roller, E.~Dinan, K.~Cho, and J.~Weston,
\newblock ``Neural text generation with unlikelihood training,''
\newblock in {\em 8th International Conference on Learning Representations, ICLR 2020}, Addis Ababa, Ethiopia, Apr. 2020.

\bibitem{ctrl}
N.~S. Keskar, B.~McCann, L.~R. Varshney, C.~Xiong, and R.~Socher,
\newblock ``{CTRL}: A conditional transformer language model for controllable generation,''
\newblock 2019.

\bibitem{guided}
M.~Ghazvininejad, X.~Shi, J.~Priyadarshi, and K.~Knight,
\newblock ``{H}afez: an interactive poetry generation system,''
\newblock in {\em Proceedings of {ACL} 2017, System Demonstrations}, Vancouver, Canada, July 2017, pp. 43--48, Association for Computational Linguistics.

\bibitem{gamma}
S.~Wu and M.~Sun,
\newblock ``Sampling with attribute-related information for controlling language models,''
\newblock {\em arXiv preprint arXiv:2205.06036}, 2022.

\bibitem{diverse}
A.~K. Vijayakumar, M.~Cogswell, R.~R. Selvaraju, Q.~Sun, S.~Lee, D.~Crandall, and D.~Batra,
\newblock ``Diverse beam search: Decoding diverse solutions from neural sequence models,''
\newblock {\em arXiv preprint arXiv:1610.02424}, 2016.

\bibitem{information}
C.~Meister, G.~Wiher, T.~Pimentel, and R.~Cotterell,
\newblock ``On the probability{--}quality paradox in language generation,''
\newblock in {\em Proceedings of the 60th Annual Meeting of the Association for Computational Linguistics (Volume 2: Short Papers)}, Dublin, Ireland, May 2022, pp. 36--45, Association for Computational Linguistics.

\bibitem{rocstories}
N.~Mostafazadeh, N.~Chambers, X.~He, D.~Parikh, D.~Batra, L.~Vanderwende, P.~Kohli, and J.~Allen,
\newblock ``A corpus and cloze evaluation for deeper understanding of commonsense stories,''
\newblock in {\em Proceedings of the 2016 Conference of the North {A}merican Chapter of the Association for Computational Linguistics: Human Language Technologies}, San Diego, California, June 2016, pp. 839--849, Association for Computational Linguistics.

\bibitem{adgen}
Z.~Shao, M.~Huang, J.~Wen, W.~Xu, and X.~Zhu,
\newblock ``Long and diverse text generation with planning-based hierarchical variational model,''
\newblock in {\em Proceedings of the 2019 Conference on Empirical Methods in Natural Language Processing and the 9th International Joint Conference on Natural Language Processing (EMNLP-IJCNLP)}, Hong Kong, China, Nov. 2019, pp. 3257--3268, Association for Computational Linguistics.

\bibitem{webnlg}
C.~Gardent, A.~Shimorina, S.~Narayan, and L.~Perez-Beltrachini,
\newblock ``The {W}eb{NLG} challenge: Generating text from {RDF} data,''
\newblock in {\em Proceedings of the 10th International Conference on Natural Language Generation}, Santiago de Compostela, Spain, Sept. 2017, pp. 124--133, Association for Computational Linguistics.

\bibitem{key}
K.~Uchimoto, S.~Sekine, and H.~Isahara,
\newblock ``Text generation from keywords,''
\newblock in {\em {COLING} 2002: The 19th International Conference on Computational Linguistics}, 2002.

\bibitem{sql}
K.~Xu, L.~Wu, Z.~Wang, Y.~Feng, and V.~Sheinin,
\newblock ``{SQL}-to-text generation with graph-to-sequence model,''
\newblock in {\em Proceedings of the 2018 Conference on Empirical Methods in Natural Language Processing}, Brussels, Belgium, Oct.-Nov. 2018, Association for Computational Linguistics.

\bibitem{table}
S.~Ma, P.~Yang, T.~Liu, P.~Li, J.~Zhou, and X.~Sun,
\newblock ``Key fact as pivot: A two-stage model for low resource table-to-text generation,''
\newblock in {\em Proceedings of the 57th Annual Meeting of the Association for Computational Linguistics}, Florence, Italy, July 2019, pp. 2047--2057, Association for Computational Linguistics.

\bibitem{constrained}
C.~Garbacea and Q.~Mei,
\newblock ``Why is constrained neural language generation particularly challenging?,''
\newblock {\em arXiv preprint arXiv:2206.05395}, 2022.

\bibitem{coverage}
A.~See, P.~J. Liu, and C.~D. Manning,
\newblock ``Get to the point: Summarization with pointer-generator networks,''
\newblock in {\em Proceedings of the 55th Annual Meeting of the Association for Computational Linguistics (Volume 1: Long Papers)}, Vancouver, Canada, July 2017, pp. 1073--1083, Association for Computational Linguistics.

\bibitem{simctg}
Y.~Su, T.~Lan, Y.~Wang, D.~Yogatama, L.~Kong, and N.~Collier,
\newblock ``A contrastive framework for neural text generation,''
\newblock {\em arXiv preprint arXiv:2202.06417}, 2022.

\bibitem{CTLoss}
S.~Jiang, R.~Zhang, S.~Vakulenko, and M.~Rijke,
\newblock ``A simple contrastive learning objective for alleviating neural text degeneration,''
\newblock {\em CoRR}, vol. abs/2205.02517, 2022.

\bibitem{forbidden-ngram}
R.~Paulus, C.~Xiong, and R.~Socher,
\newblock ``A deep reinforced model for abstractive summarization,''
\newblock in {\em 6th International Conference on Learning Representations, ICLR 2018}, Vancouver, Canada, Apr. 2018.

\bibitem{quality-diversity-tradeoff}
Z.~Hugh, D.~Daniel, I.~Daphne, and N.~Arvind,
\newblock ``Trading off diversity and quality in natural language generation,''
\newblock in {\em Proceedings of the Workshop on Human Evaluation of NLP Systems (HumEval)}, Online, Apr. 2021, pp. 25--33, Association for Computational Linguistics.

\bibitem{strategy-task}
G.~Wiher, C.~Meister, and R.~Cotterell,
\newblock ``On decoding strategies for neural text generators,''
\newblock {\em arXiv preprint arXiv:2203.15721}, 03 2022.

\bibitem{oversampling}
I.~Daphne, K.~Reno, S.~Jo{\~a}o, K.~Maria, and C.~Chris,
\newblock ``Comparison of diverse decoding methods from conditional language models,''
\newblock in {\em Proceedings of the 57th Annual Meeting of the Association for Computational Linguistics}, Florence, Italy, July 2019, Association for Computational Linguistics.

\bibitem{uid-cognitive}
J.~Wei, C.~Meister, and R.~Cotterell,
\newblock ``A cognitive regularizer for language modeling,''
\newblock in {\em Proceedings of the 59th Annual Meeting of the Association for Computational Linguistics and the 11th International Joint Conference on Natural Language Processing (Volume 1: Long Papers)}, Online, Aug. 2021, pp. 5191--5202, Association for Computational Linguistics.

\bibitem{uid-beam}
C.~Meister, R.~Cotterell, and T.~Vieira,
\newblock ``If beam search is the answer, what was the question?,''
\newblock in {\em Proceedings of the 2020 Conference on Empirical Methods in Natural Language Processing (EMNLP)}, Online, Nov. 2020, pp. 2173--2185, Association for Computational Linguistics.

\bibitem{speaking-human}
A.~F. Frank and T.~F. Jaeger,
\newblock ``Speaking rationally: Uniform information density as an optimal strategy for language production,''
\newblock in {\em In Cogsci}, 2008.

\bibitem{ROUGE}
C.~Lin,
\newblock ``{ROUGE}: A package for automatic evaluation of summaries,''
\newblock in {\em Text Summarization Branches Out}, Barcelona, Spain, July 2004, pp. 74--81, Association for Computational Linguistics.

\bibitem{dist}
J.~Li, M.~Galley, C.~Brockett, J.~Gao, and B.~Dolan,
\newblock ``A diversity-promoting objective function for neural conversation models,''
\newblock in {\em Proceedings of the 2016 Conference of the North {A}merican Chapter of the Association for Computational Linguistics: Human Language Technologies}, San Diego, California, June 2016, pp. 110--119, Association for Computational Linguistics.

\bibitem{ppl}
D.~Jurafsky and J.~H. Martin,
\newblock {\em Speech and Language Processing},
\newblock Pearson Education India, 2000.

\bibitem{t5}
C.~Raffel, N.~Shazeer, A.~Roberts, K.~Lee, S.~Narang, M.~Matena, Y.~Zhou, W.~Li, and P.~J. Liu.,
\newblock ``Exploring the limits of transfer learning with a unified text-to-text transformer,''
\newblock vol. 21, no. 1, jan 2020.

\bibitem{gpt2}
A.~Radford, J.~Wu, R.~Child, D.~Luan, D.~Amodei, and I.~Sutskever,
\newblock ``Language models are unsupervised multitask learners,''
\newblock {\em OpenAI blog}, vol. 1, no. 8, pp. 9, 2019.

\end{thebibliography}

\vfill\pagebreak

\appendix

\onecolumn

\section{More Cases for CommonGEN}

\label{sec:app1}

\begin{table}[htb]
	\centering
	\label{table:app1}
	\caption{More cases generated by our fine-tuned model on CommonGEN.}
	\begin{tabular}{cc}
	   \hline
          Input & \pbox{12cm}{climb talk wall} \\\hline
          Reference & \pbox{12cm}{The woman thats talking teaches her students how to climb the wall.} \\\hline
          Beam. & \pbox{12cm}{A man is climbing a wall and talking to someone.} \\
          Gamma. & \pbox{12cm}{A young boy climbs the wall in conversation with his father.}\\
          Greedy & \pbox{12cm}{A man is climbing a wall and talking to a man.}\\
          Temp. & \pbox{12cm}{A man talks about climbing the wall.}\\
          Topk. & \pbox{12cm}{A group of people are climbing a wall to talk to each other.}\\
          IFDID & \pbox{12cm}{A young boy climbs the wall and talks to someone.}\\
          IFDID-SIMI. & \pbox{12cm}{A young boy climbs the wall to talk about how much it costs him to live on his own} \\
          \hline
    \end{tabular}
\end{table}

\section{MORE CASES FOR ROCSTORIES}

\label{sec:app2}

\begin{longtable}{cc}
	\caption{Case study on RocStories, which contains repetitive results to be analyzed. \textcolor{red}{The repetitive part} is highlighted.}
	\label{table:app2}\\
	   \specialrule{0.1em}{5pt}{5pt}
          Input & \pbox{12cm}{Skipping School} \\\specialrule{0.1em}{5pt}{5pt}
          Reference & \pbox{12cm}{Princeton had overheard his mom on the phone with bill collectors. Until that moment, he didn't know how hard his mom was struggling. His principal alerted her to his absenteeism at school. When he came home from school, she questioned him about it. It was then he gave her money he had saved from the time job he found.} \\\specialrule{0.1em}{5pt}{5pt}
          Topp. & \pbox{12cm}{Tom was at school. He was having trouble \textcolor{red}{getting into class}. He decided to skip school. Tom was able to \textcolor{red}{get into class}. He was able to get into class.} \\\specialrule{0em}{5pt}{5pt}
          Gamma. & \pbox{12cm}{Charles didn't know how to snowboard. He wasn't able to learn the lesson at first. Then he decided to try ski jumping. After about an hour, Charles got good enough for his class. Now Charles has a fun day that no one else can forget.}\\\specialrule{0em}{5pt}{5pt}
          Greedy & \pbox{12cm}{I was at a ski resort. I \textcolor{red}{was going to} the top of the mountain. I \textcolor{red}{was going to} ski down the slope. I \textcolor{red}{was going to} drop my ski poles. I fell off the mountain. I had to go home. I was happy that I had skipped school. I hope I never skied again.}\\\specialrule{0em}{5pt}{5pt}
          Temp. & \pbox{12cm}{Tom was at the park. He was taking a break from school. He decided to skip school. Tom \textcolor{red}{got into trouble} with the school principal. Tom was suspended for the rest of the day. Tom did not \textcolor{red}{get into trouble} again. Tom was glad he skipped school.}\\\specialrule{0em}{5pt}{5pt}
          Topk. & \pbox{12cm}{Tom was at his school. He had to skip school. His parents told him he would have to go back. Tom didn't listen. He skipped school and got into trouble. Tom had to go back to school. His parents had to pay for his suspension. Tom had to go back to school and never went back. Tom \textcolor{red}{was glad he skipped school}. He \textcolor{red}{was glad he skipped school and got into better shape}. Tom \textcolor{red}{was glad he skipped school and got into better shape.}}\\\specialrule{0em}{5pt}{5pt}
          IFDID & \pbox{12cm}{Ann was having trouble getting her classes in order. She needed to start school early so she could get some sleep. After going through the website, she went to a library. There, she read about various resources for learning new skills. When it was time to head home, she decided not to go back to school. Instead, she chose to skip class on Thursday night instead. Now, Ann is able to focus more and better at her upcoming math test.}\\\specialrule{0em}{5pt}{5pt}
          IFDID-SIMI. & \pbox{12cm}{Today we went to the beach. We saw that there were many fish in our boat and rowboat! As soon as I got off of shore, my brother suggested taking a swim at sea school. My parents agreed so they took us back home after class for lunch later. After dinner with family members from School Day Skipping We returned home empty handed because not enough money was left on board. By day's end however one student made \$100 just by jumping out into water alone. Now since she does very well online teacher comes around here every week gives her extra cash too} \\\specialrule{0.1em}{5pt}{5pt}
\end{longtable}

\begin{longtable}{cc}
	\caption{Case study on RocStories. \textcolor{red}{The repetitive part} is highlighted.}
	\label{table:app3}\\
	   \specialrule{0.1em}{5pt}{5pt}
          Input & \pbox{12cm}{Dancing} \\\specialrule{0.1em}{5pt}{5pt}
          Reference & \pbox{12cm}{I used to be very bad at dancing. I wanted to be very good at dancing. I practiced dancing in my room for years and years. I even took lessons from a professional. Now I am a very good dancer.} \\\specialrule{0.1em}{5pt}{5pt}
          Topp. & \pbox{12cm}{We went to the dance studio to learn how to dance. We practiced all day. We practiced all night. We danced all night. We had a great time.} \\\specialrule{0em}{5pt}{5pt}
          Gamma. & \pbox{12cm}{Tim always loved dance. He started taking lessons and tried out for the job. His first night he didn't even know how to dance. It got very challenging, but Tim finally learned to do it. Now Tim's favorite type of dancing is called snooze dance. Tim still loves his new hobby! Tim will try again next time he wants to learn more about dance.}\\\specialrule{0em}{5pt}{5pt}
          Greedy & \pbox{12cm}{I was dancing with my friends. \textcolor{red}{I was nervous}. \textcolor{red}{I was nervous} to be in front of a crowd.}\\\specialrule{0em}{5pt}{5pt}
          Temp. & \pbox{12cm}{The kids were dancing. They were making a big splash. They got all the attention. They won the competition. \textcolor{red}{They were very proud of} themselves. \textcolor{red}{They were proud of their dancing skills}. \textcolor{red}{They were proud of their dancing skills} and their parents.}\\\specialrule{0em}{5pt}{5pt}
          Topk. & \pbox{12cm}{The kids were dancing. They were having a great time. They had a lot of fun. They went home and played for hours. They were happy that they had fun. They had a great time at home. They were proud of their dancing friends.}\\\specialrule{0em}{5pt}{5pt}
          IFDID & \pbox{12cm}{James had always wanted to learn how to dance. He decided he would try out for the school's dancing team. When they announced their selection, James was nervous. They told him that he needed a little practice time before getting into shape. James ended up becoming very good at his skills and became an instructor! Now James is happy that he has learned how to dance. His friends all agree that it's been a great year of learning about dance. Hopefully this week will be no different as James knows he'll get in better shape soon.}\\\specialrule{0em}{5pt}{5pt}
          IFDID-SIMI. & \pbox{12cm}{Although they were young, both girls loved playing with swords. They began to practice every day at school and then home for three months straight. One morning when their teacher took them out of class he announced that his daughter had gone missing! She was found running through the woods just before dawn on her way back from work. Her hair looked very sad but she knew it wasn't good news - yet! After looking into each other's eyes all night long there never seemed any cause for alarm. Because nothing could stop one girl now until another finally finds him. Maybe once more my sweet brother will find me soon enough Dancer While learning how to dance is something I always wanted as a child When i got older this means love More talking than fighting My favorite thing about dancing though But last time we fought together today So getting readyFor weeks things started coming quickSo decided doing everything really fast worked away From proportue shouldn't mean Anymore Before leaving Even if you're wrong What does matter To end All relationships become permanent Good luck buying confidence Everytime someone tells You make sureYou don't mess up Your parents tell Me No Make no mistakeDon't let anyone else bring down your dreams Realize reality Then Get along With Them Right Now Despite being born During college Today became Using Between Class Two boys hate math Why did challeng These two.} \\\specialrule{0.1em}{5pt}{5pt}
\end{longtable}

\section{MORE CASES FOR ADGEN}
\label{sec:app3}
\begin{longtable}{cc}
	\caption{Case study on AdGen. \textcolor{blue}{The repetitive part} is highlighted. The part of the output text that successfully covers the input is \underline{underlined}.\label{long}}
	\label{table:app4}\\
	   \specialrule{0.1em}{5pt}{5pt}
          Input & \pbox{12cm}{\begin{CJK}{UTF8}{gbsn}[类型: 裤], [颜色: 绿色], [风格: 复古], [风格: 知性], [图案: 花色], [图案: 复古], [图案: 线条], [图案: 印花], [裤长: 长裤], [裤型: 阔腿裤]\end{CJK}} \\\specialrule{0em}{5pt}{5pt}
           & \pbox{12cm}{[Type: Pants], [Color: Green], [Style: Vintage], [Style: intellectual], [Pattern: Floral], [Pattern: Vintage], [Pattern: Line], [Pattern: Print], [Pant Length: Long Pants], [Pant Type: Wide-Leg]} \\\specialrule{0.1em}{5pt}{5pt}
          Reference & \pbox{12cm}{\begin{CJK}{UTF8}{gbsn}\underline{绿色}的基调打造的这款花色\underline{长裤}，整体采用了垂坠感十足的雪纺纱质地，结合精美的大\underline{印花}点缀，带来\underline{复古}且\underline{知性}的\underline{阔腿裤}剪裁，是非常实穿的组合。设计师为这款裤子做了长款的剪裁，一直盖住脚背，充分拉长了腿部的\underline{线条}感。\end{CJK}} \\\specialrule{0em}{5pt}{5pt}
           & \pbox{12cm}{Created in a \underline{green} tone, these floral \underline{long pants} have an overall draped chiffon gauze texture combined with a beautiful large \underline{print} embellishment for a \underline{vintage} and \underline{intellectual} \underline{wide-leg} silhouette, a very wearable combination. The designer has made these pants in a long cut that goes all the way down to cover the instep, fully elongating the legs.} \\\specialrule{0.1em}{5pt}{5pt}
          Beam. & \pbox{12cm}{\begin{CJK}{UTF8}{gbsn}这款\underline{阔腿裤}采用了\underline{复古}的\underline{印花}图案设计，在视觉上增添了几分\underline{知性}的气息，让整体看起来更加的\textcolor{blue}{\underline{知性}\underline{知性}}。宽松的\underline{阔腿}版型，修饰\textcolor{blue}{腿部\underline{线条}}的同时，还能够提\textcolor{blue}{腿部的\underline{线条}}感。\end{CJK}} \\\specialrule{0em}{5pt}{5pt}
           & \pbox{12cm}{Designed with a \underline{vintage} print, these \underline{wide-leg pants} \textcolor{blue}{add a touch of intellectuality to the visuals}, \textcolor{blue}{giving the overall look a more intellectual and \underline{intellectual} feel}. The relaxed, \underline{wide-leg} fit trims the legs while also enhancing them.} \\\specialrule{0.1em}{5pt}{5pt}
          Beam${_{without}}$ & \pbox{12cm}{\begin{CJK}{UTF8}{gbsn}这款\underline{阔腿裤}采用了\underline{复古}的\underline{印花}图案，在视觉上增添了几分\textcolor{blue}{\underline{复古}的韵味}，同时也增添了几分\textcolor{blue}{\underline{复古}的韵味}。这款\underline{阔腿裤}采用了\underline{复古}的\underline{印花}图案，在视觉上增添了几分\textcolor{blue}{\underline{复古}的韵味}，同时也\textcolor{blue}{增添了几分\underline{复古}的韵味}。\end{CJK}} \\\specialrule{0em}{5pt}{5pt}
           & \pbox{12cm}{Featuring a \underline{vintage} \underline{print}, these \underline{wide-legged pants} \textcolor{blue}{add a touch of \underline{vintage} charm to the visuals}, while also adding a touch of retro charm. Featuring a \underline{vintage} \underline{print}, these \underline{wide-legged pants}\textcolor{blue}{ add a touch of \underline{vintage} charm to the visuals}, while also \textcolor{blue}{adding a touch of retro charm.}} \\\specialrule{0.1em}{5pt}{5pt}
          Topk. & \pbox{12cm}{\begin{CJK}{UTF8}{gbsn}这款\underline{阔腿裤}采用了经典的\underline{印花}设计，穿着舒适，不易褪色，不易变形。而且还能修饰腿型，穿着更显时髦。而且，这款\underline{阔腿裤}的设计也非常简单。经典的\underline{印花}图案，让你的腿部\underline{线条}感受不断变得轻松自由，更显\underline{知性}。\end{CJK}} \\\specialrule{0em}{5pt}{5pt}
           & \pbox{12cm}{These \underline{wide-legged pants} feature a classic \underline{printed} design that is comfortable to wear and not easily faded or distorted. It also slims the leg shape and makes it more fashionable to wear. Moreover, the design of these \underline{wide-legged pants} is very simple. With the classic \underline{print}, your legs feel constantly relaxed and free and more \underline{intellectual}.} \\\specialrule{0.1em}{5pt}{5pt}
           Nucleus. & \pbox{12cm}{\begin{CJK}{UTF8}{gbsn}这款休闲\textcolor{blue}{阔腿\underline{阔腿裤}}，选用\underline{绿色}的花色，搭配\underline{复古}的花色，更具\underline{知性}感，穿着舒适。而且裤脚采用的是\underline{复古}的\underline{印花}，让裤子看起来更加的时尚大方，而且还带有很好的视觉效果，同时还具有时髦感。\end{CJK}} \\\specialrule{0em}{5pt}{5pt}
           & \pbox{12cm}{These casual \textcolor{blue}{wide-legged \underline{wide-legged pants}} in \underline{green} with a \underline{vintage} floral colour are more \underline{intellectual} and comfortable to wear. The trouser legs are in a vintage \underline{print}, giving them a stylish and generous look with a great visual effect and a chic feel at the same time.} \\\specialrule{0.1em}{5pt}{5pt}
          Gamma. & \pbox{12cm}{\begin{CJK}{UTF8}{gbsn}衣身的\underline{线条}显得格外清爽，穿上它就更加有范儿了。精致的\underline{绿色}拼接花边图案点缀在腿部，打造出浪漫梦幻的视觉效果。深灰色的\underline{印花}撞色的纹理提升了时尚感，突显品牌的新颖形象。不仅如此，裙身的细节也极为有趣，给人很好的视觉冲击力。\end{CJK}} \\\specialrule{0em}{5pt}{5pt}
           & \pbox{12cm}{The \underline{lines} of the bodice look extra crisp, making it even more stylish to wear. A delicate \underline{green} patchwork lace pattern embellishes the legs, creating a romantic and dreamy visual effect. The texture of the dark grey \underline{print} clashes with the colour enhances the sense of style and highlights the brand's fresh image. Not only that, but the details on the skirt are extremely interesting and give a great visual impact.} \\\specialrule{0.1em}{5pt}{5pt}
          IFDID & \pbox{12cm}{\begin{CJK}{UTF8}{gbsn}这款\underline{阔腿裤}非常百搭，比较经典的版型设计，更加\underline{知性}和浪漫，看起来帅气干练又优雅。\underline{绿色}的花色增添几分温暖之感，而且还可以让你轻松驾驭不同风格的运动鞋。休闲短裤的设计，带来更多的视觉冲击，带来更加满满的\underline{知性}气息。\end{CJK}} \\\specialrule{0em}{5pt}{5pt}
           & \pbox{12cm}{These \underline{wide-legged pants} are very versatile, with a more classic fit design that is more \underline{intellectual} and romantic, looking handsome and competent yet elegant. The \underline{green} floral colour adds a bit of warmth and also allows you to easily manage different styles of sneakers. The design of the casual shorts brings more visual impact and a more full of \underline{intellectuality}.} \\\specialrule{0.1em}{5pt}{5pt}
          IFDID-SIMI & \pbox{12cm}{\begin{CJK}{UTF8}{gbsn}腰部的纹理紧致美观，有效增强孩子气息和\underline{知性}简洁的色彩搭配，提升层次感也是帅气吸睛了不少。经典的绣花图案将生态染得淋漓尽致，百搭又实用。整体的立体\underline{印花}更是满足全家所需要的哦。休闲裙摆撞色的纽扣蕾丝点缀，就特别服帖时髦减龄。\end{CJK}} \\\specialrule{0em}{5pt}{5pt}
           & \pbox{12cm}{The texture of the waist is tight and beautiful, effectively enhancing the child's breath and the \underline{intellectual} simplicity of the colour scheme, enhancing the layering is also handsome and eye-catching. The classic embroidered pattern is ecologically dyed to perfection, versatile and practical. The overall three-dimensional \underline{print} is a great way to meet the needs of the whole family. The casual hemline is embellished with button lace, which is particularly flattering and chic.} \\\specialrule{0.1em}{5pt}{5pt}
\end{longtable}

\section{HUMAN EVALUATION GUIDANCE}
\label{sec:app-human}

In this section, we give the sheet templates that are used to human evaluation.

\subsection{Commonsense generation}

We evaluate CommonGEN, the commonsense generation task, with questions as follows:

\noindent {\bf 1. [fluency] Is the sentence well-formed?}

\begin{itemize}
\setlength{\itemsep}{0pt}
\setlength{\parsep}{0pt}
\setlength{\parskip}{0pt}
\item score \textbf{1}: Yes. The statement is well-formed, fluent and is grammatically correct.
\item score \textbf{2}: Somewhat. The statement is understandable but have some grammar errors.
\item score \textbf{3}: No. The statement is neither well-formed nor fluent and includes grammar errors.
\end{itemize}

\noindent {\bf 2. [diversity] Is there a wide range of words and phrases in the sentence?}

\begin{itemize}
\setlength{\itemsep}{0pt}
\setlength{\parsep}{0pt}
\setlength{\parskip}{0pt}
\item score \textbf{1}: Yes. The sentence not only contains concepts above, but has colorful words and expressions as well.
\item score \textbf{2}: Somewhat. The sentence has extra words but not enough.
\item score \textbf{3}: No. The sentence is dull and boring. It only covers the concepts.
\end{itemize}

\noindent {\bf 3. [faithfulness] Does the sentence include the given concepts meaningfully?}

\begin{itemize}
\setlength{\itemsep}{0pt}
\setlength{\parsep}{0pt}
\setlength{\parskip}{0pt}
\item score \textbf{1}: Yes. The sentence meaningfully includes all of the concepts.
\item score \textbf{2}: Somewhat. Some, but not all, of the concepts are included in the statement in a meaningful way. Alternatively, the statement contains all concepts, but some of them are meaningless or improperly combined.
\item score \textbf{3}: No. The statement does not incorporate any significant notions.
\end{itemize}

\noindent {\bf 4. [overall] Considering your answer to 1,2 and 3, does the sentence meaningfully combine all of the concepts and have high-diersity expression?}

\begin{itemize}
\setlength{\itemsep}{0pt}
\setlength{\parsep}{0pt}
\setlength{\parskip}{0pt}
\item score \textbf{1}: Yes. The sentence meaningfully combines all the concepts into a sentence with high-diversity.
\item score \textbf{2}: Somewhat. In terms of the preceding questions, the sentence appears to be correct.
\item score \textbf{3}: No. The sentence is incomprehensible, uninteresting, or fails to effectively incorporate all of the elements into a sentence.
\end{itemize}

\subsection{Story generation}

We evaluate RocStories, the story generation task, with questions below:

\noindent {\bf 1. [grammar] Is the story written in a grammatically correct way?}

\begin{itemize}
\setlength{\itemsep}{0pt}
\setlength{\parsep}{0pt}
\setlength{\parskip}{0pt}
\item score \textbf{1}: Yes. It is entirely or mostly grammatically correct, with no or few grammatically errors.
\item score \textbf{2}: Somewhat. It is partially grammatically correct, with some grammatical errors.
\item score \textbf{3}: No. It is mostly not grammatically correct, with many grammatical errors.
\end{itemize}

\noindent {\bf 2. [fluency] Is the story written in a fluent and understandable way?}

\begin{itemize}
\setlength{\itemsep}{0pt}
\setlength{\parsep}{0pt}
\setlength{\parskip}{0pt}
\item score \textbf{1}: Yes. It is totally or mostly fluent and intelligible.
\item score \textbf{2}: Somewhat. It is somewhat fluent and intelligible, but it reads awkwardly.
\item score \textbf{3}: No. It is mainly poorly written and hard to understand.
\end{itemize}

\noindent {\bf 3. [faithfulness] Does the story stay on topic?}

\begin{itemize}
\setlength{\itemsep}{0pt}
\setlength{\parsep}{0pt}
\setlength{\parskip}{0pt}
\item score \textbf{1}: Yes. The story totally stays on topic.
\item score \textbf{2}: Somewhat. The story reads a little off topic.
\item score \textbf{3}: No. It is mostly off topic.
\end{itemize}

\noindent {\bf 4. [interestingness] Is the story written in an interesting way?}

\begin{itemize}
\setlength{\itemsep}{0pt}
\setlength{\parsep}{0pt}
\setlength{\parskip}{0pt}
\item score \textbf{1}: Very interesting: The story has themes, characters, and dialog that keeping you reading wanting to show it to a friend.
\item score \textbf{2}: Somewhat interesting: The story has themes, characters, dialog, and/or a writing style that piques your attention.
\item score \textbf{3}: Mildly interesting: There are fascinating moments but the plot isn't very memorable.
\item score \textbf{4}: Not very interesting: You complete the story but have no recollection of anything special about it.
\item score \textbf{5}: Not at all interesting: You don't want to finish reading the story since it's uninteresting. It is either dull or unimaginative.
\end{itemize}

\noindent {\bf 5. [overall] Considering the above questions, overall, what's the quality of the story?}

\begin{itemize}
\setlength{\itemsep}{0pt}
\setlength{\parsep}{0pt}
\setlength{\parskip}{0pt}
\item score \textbf{1}: The overall quality is high.
\item score \textbf{2}: The overall quality is okay.
\item score \textbf{3}: The overall quality is low.
\end{itemize}

\subsection{Specific style text generation}

In this section, we give the evaluation sheets on AdGen task:

\noindent {\bf 1. [grammar] Do the generated sentences contain Chinese grammatical errors?}

\begin{itemize}
\setlength{\itemsep}{0pt}
\setlength{\parsep}{0pt}
\setlength{\parskip}{0pt}
\item score \textbf{1}: Yes. The sentence includes grammatical errors and they affect the understanding of the sentence.
\item score \textbf{2}: Somewhat. The sentence includes grammatical errors but they do not affect understanding.
\item score \textbf{3}: No. Fluent sentences, and no grammatical errors.
\end{itemize}

\noindent \textbf{2. [faithfulness] Is the generated sentence closely related to the given keywords, logically coherent and well-organized?}

\begin{itemize}
\setlength{\itemsep}{0pt}
\setlength{\parsep}{0pt}
\setlength{\parskip}{0pt}
\item score \textbf{1}: Yes. Generated sentences can cover all descriptive words in the input text, logically coherent and well-organized.
\item score \textbf{2}: Somewhat. The generated sentences cover some words in the input text and there is some information that is not covered.
\item score \textbf{3}: No. The generated sentences contain too much irrelevant information or basically do not cover the information in the input text.
\end{itemize}

\noindent \textbf{3. [diversity] Is the amount of words used in the generated text large enough and the Chinese expressions rich enough?}

\begin{itemize}
\setlength{\itemsep}{0pt}
\setlength{\parsep}{0pt}
\setlength{\parskip}{0pt}
\item score \textbf{1}: Yes. The generated text is rich in words and free of repetitive words.
\item score \textbf{2}: Somewhat. The generated text contains no repetitive phrases/subsentences, but it is dull.
\item score \textbf{3}: No. The generated text has single syntax and too many repetitive phrases/subsentences.
\end{itemize}

\noindent \textbf{4. [overall] Combining the three questions above, do the generated sentences cover all the information in the input text with a variety of words and no grammatical errors?}

\begin{itemize}
\setlength{\itemsep}{0pt}
\setlength{\parsep}{0pt}
\setlength{\parskip}{0pt}
\item score \textbf{1}: Yes. The generated sentence covers all or most of the information given in the input text, with a rich vocabulary and no obvious grammatical errors.
\item score \textbf{2}: Somewhat. The generated sentence covers only part of the information in the input text, with single words, simple sentences, or a few grammatical errors.
\item score \textbf{3}: No. The generated sentence does not cover all the information in the input text, which is too simple, or has obvious grammatical errors.
\end{itemize}

\end{document}